\pdfoutput=1
\documentclass[11pt]{article}
\usepackage[final]{acl}

\usepackage{silence}
\usepackage{newtxtext}
\usepackage{newtxmath}
\usepackage{inconsolata}
\usepackage{microtype}
\usepackage[T1]{fontenc}
\usepackage[utf8]{inputenc}

\usepackage{balance}
\usepackage{booktabs}
\usepackage{csquotes}
\usepackage[
    byauthor,
    separator={;\ }
]{credits}
\usepackage{float}
\usepackage{graphicx}
\usepackage{lipsum}
\usepackage{soul}
\usepackage{subcaption}
\usepackage{tabularx}
\usepackage{longtable}
\newcommand{\Manar}[1]{}

\newcommand{\Simeon}[1]{}

\newcommand{\Sina}[1]{}

\newcommand{\annolabel}[1]{\textit{#1}}
\newcommand{\variable}[1]{\textit{#1}}

\title{Testing Associative Creativity in Humans and Multimodal LLMs\\
with Difficult-to-Name Visual Stimuli}

\author{%
    Manar Ali\textsuperscript{} \thanks{Equal Contribution}\:
    Lisa Gottschalk\textsuperscript{\footnotemark[1]} \:
    Simeon Junker\textsuperscript{} \: \\
    \bf\large
    Lisa Hudalla \textsuperscript{}
    Alon Fishman \textsuperscript{} \: 
    Joana Cholin \textsuperscript{}
    Stefan Kopp \textsuperscript{} \\
    \bf\large
    Hendrik Buschmeier\textsuperscript{} \:
    Sina Zarrieß\textsuperscript{}
\\
    CRC 1646 ‘Linguistic Creativity in Communication’
\\ 
    Bielefeld University, Bielefeld, Germany\\
       \textbf{Correspondence:} \texttt{\{manar.ali|lisa.gottschalk\}@uni-bielefeld.de}\\
    }

\credit{Manar Ali}{
    Conceptualization,
    Data curation,
    Formal analysis,
    Investigation,
    Methodology,
    Resources,
    Software,
    Visualization,
    Writing -- original draft,
    Writing -- review \& editing}
\credit{Lisa Gottschalk}{
    Conceptualization,
    Data curation,
    Formal analysis,
    Investigation,
    Methodology,
    Resources,
    Software,
    Visualization,
    Writing -- original draft,
    Writing -- review \& editing}
\credit{Simeon Junker}{
    Conceptualization,
    Methodology,
    Software,
    Data curation,
    Writing -- original draft,
    Writing -- review \& editing}
\credit{Alon Fishman}{
    Resources,
    Writing -- review \& editing}
\credit{Lisa Hudalla}{
    Software}
\credit{Joana Cholin}{
    Supervision}
\credit{Stefan Kopp}{
    Supervision,
    Writing -- review \& editing}
\credit{Hendrik Buschmeier}{
    Conceptualization,
    Supervision,
    Writing -- original draft,
    Writing -- review \& editing}
\credit{Sina Zarrieß}{
    Conceptualization,
    Supervision,
    Writing -- original draft,
    Writing -- review \& editing}

\begin{document}
\maketitle

\begin{abstract}
We investigate associative creativity in descriptions of abstract visual stimuli by humans and multimodal large language models (MLLMs). 
While conceptual association is central to human communication, it remains unclear how humans and models adapt when visual stimuli become increasingly difficult to map onto familiar semantic concepts. 
We evaluate several state-of-the-art MLLMs and human participants along two dimensions of creativity, originality and effectiveness, using two datasets with controlled manipulations of difficulty. 
\textit{SilCC} varies naming difficulty through progressively more complex combinations of abstract shapes, and \textit{PixCC} varies difficulty through increasing levels of random visual noise. 
Results show that humans produce more diverse associative descriptions than MLLMs, including greater semantic distance, higher entropy, and more novel expressions relative to reference lexica. 
As difficulty increases, effectiveness declines in \textit{SilCC}, and humans increasingly shift from holistic associations to descriptions of low-level visual properties in \textit{PixCC}.%
\footnote{We will release all data and code via Zenodo upon acceptance of the paper.}
\end{abstract}

\section{Introduction}
\label{sec:introduction}

Humans use conceptual associations to describe novel experiences, overcome communicative challenges, and convey meaning even when the precise vocabulary is unavailable. For example, a toddler encountering an airplane for the first time might describe it as a \enquote{bird}. Such conceptual mapping reflects a fundamental mechanism of human cognition and has been linked to domain-specific performance across a range of tasks \citep{Beaty2023}. \Sina{can we include a sentence here that connects conceptual association to (linguistic) creativity? that would make the transition to the next paragraph more smooth  }

Recent research on multimodal large language models (MLLMs) has begun to examine and benchmark their capacity to produce creative descriptions of different kinds of input images (see following section).
    \Sina{maybe we can phrase this a bit differently here? research on MLLMs has begun to examine their capacity to generate descriptions of abstract images, but has not systematically framed and evaluated this as a creative capability ... it is unclear how  systems (and humans) trade off originality and effectiveness of their descriptions ... this trade off is particularly interesting to look at under conditions of increase communicative difficulty, i.e., when associations are less conventionalized and originality + effectiveness might be more difficult to achieve at the same time ... we could also stress a bit more here that existing studies often pick a particular type of abstract image(e.g., tangram, color grid , ...) but do not compare between different ways of introducing communicative difficulty (e.g., noise vs. complexity) ...}
However, it remains unclear how such systems respond to stimuli of increasing difficulty, particularly in cases when associations become less straightforward due to increased visual complexity, reduced resemblance to familiar objects, or reduced ‘degree of objecthood’ \citep{feldman2003what}.
In principle, several outcomes are conceivable: 
Systems with strong associative capabilities may continue to connect visual inputs to meaningful semantic concepts despite increased difficulty. 
Alternatively, they may shift their descriptive strategies, relying less on higher-level associations and more on low-level, objective properties of the stimulus. 
Finally, their ability to generate meaningful descriptions may deteriorate altogether. To date, these possibilities have not been systematically examined in either humans or generative systems.

\begin{figure}
\centering
    \includegraphics[width=0.2\linewidth]{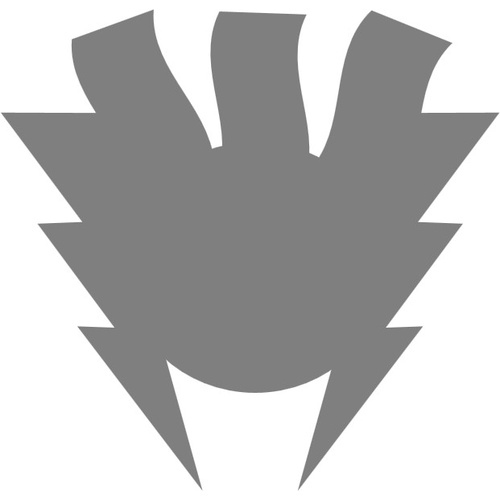}
    \hspace{1em}
    \includegraphics[width=0.2\linewidth]{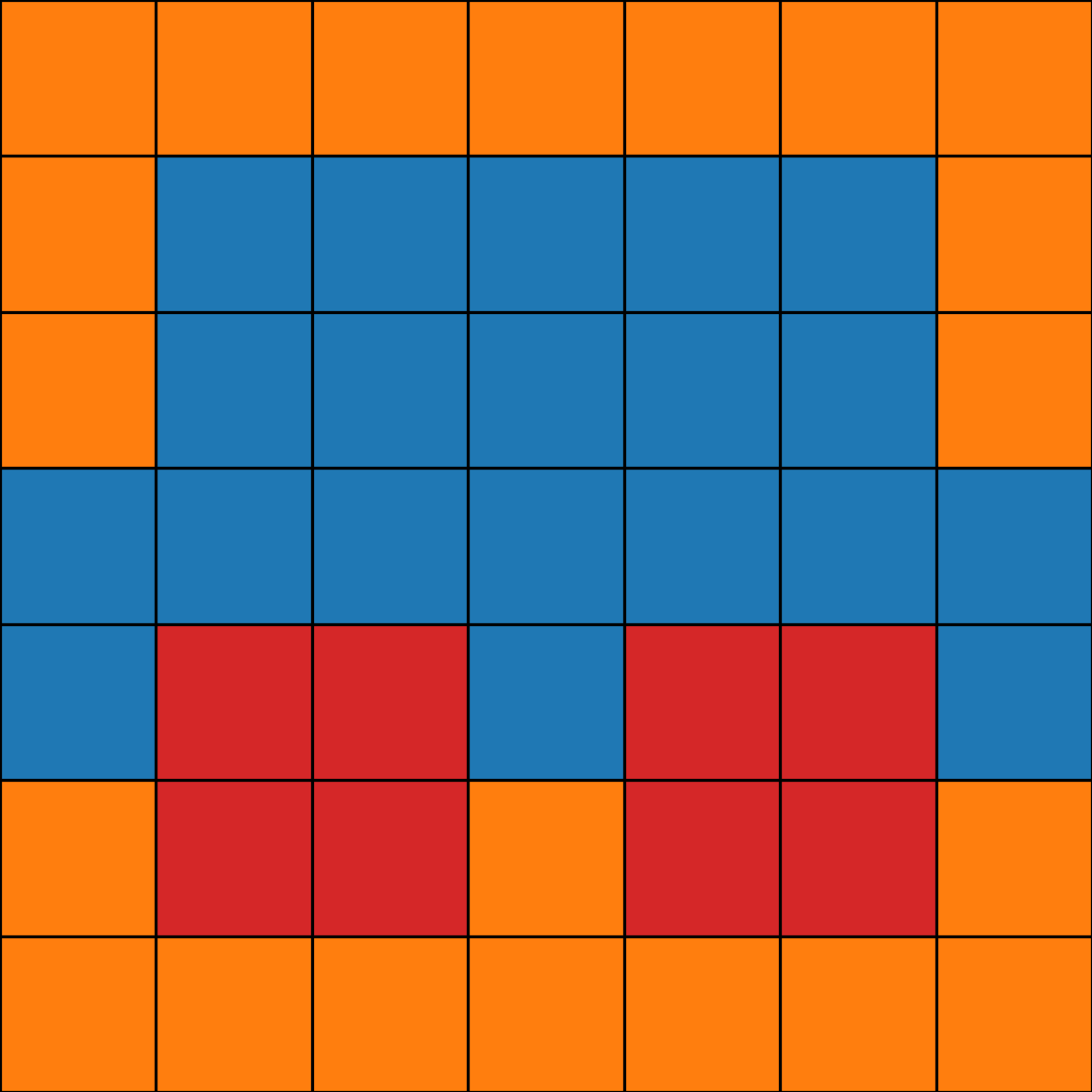}
    \caption{%
        Example items from the two datasets, SilCC (left) from the highest difficulty level and PixCC (right) in clean condition. Human descriptions reflect associative naming (e.g., ``Krabbenraumschiff'' (crab spaceship), ``Legohaarexplosion'' (lego hair explosion) and ``Teufel'' (devil) for SilCC and ``sofa and two cushions'', ``car'', ``spaceship'' for PixCC.
    }
\label{fig:leading_example}
\end{figure}

In this work, we examine the associative abilities of several state-of-the-art MLLMs using established criteria for \textit{creativity}, namely \textit{originality} and \textit{effectiveness} \citep{runco2012standard}.
For this, we collect and annotate two datasets of abstract shapes with controlled manipulations of naming difficulty.
In \textit{SilCC}, difficulty is varied through the complexity of abstract shape combinations; in \textit{PixCC}, it is varied through increasing levels of random visual noise. 
Using these datasets, we analyze how naming behavior changes as difficulty increases and compare human and MLLM performance in producing original and effective descriptions.
We assess \textit{effectiveness} by testing whether descriptions adequately capture the visual input, and \textit{originality} through measures of semantic diversity and naming variation. In \textit{PixCC}, we also account for strategies that circumvent associative descriptions when noise weakens the link between visual stimuli and familiar semantic concepts.

Our work contributes to a growing area of research on creativity in language models.
However, as noted by \citet{Nakajima2026}, many existing studies that apply \enquote{creativity tests} to language models focus only on divergent abilities. 
Our image-description setting naturally combines originality and effectiveness in a single setting: originality is required because the inputs lack conventionalized names, and effectiveness is required because the descriptions must faithfully represent the inputs in a way that allows a potential listener to identify the image (see Figure~\ref{fig:leading_example} for an example).

Our results show that, across datasets and models, humans produce more diverse associative responses than MLLMs, as reflected in greater semantic distance, higher entropy, and a stronger tendency to generate novel expressions not included in ground-truth reference lexica. 
In \textit{SilCC}, effectiveness declines for both humans and models as visual complexity and naming difficulty increase. In \textit{PixCC}, we observe that humans shift toward alternative communicative strategies under high noise, with descriptions focusing on low-level visual properties rather than providing high-level interpretations.

\section{Background}
\label{sec:background}

\paragraph{Associative Creativity}

Creativity is commonly defined as the combination of \textit{originality} and \textit{effectiveness} \citep{runco2012standard, vogel2026linguistic}. 
The ability to generate ideas that satisfy both criteria is often linked \textit{convergent} and \textit{divergent} thinking \citep{Benedek2012}.
In this work, we focus in particular on associative patterns linking abstract figures to possible descriptions. 
Here, \textit{association} refers to the process of linking concepts in memory \citep{Mednick1962}. Associations can be spontaneous (free) or controlled (goal-directed) and are often evaluated using semantic distance to capture how far people move through semantic space during memory search \citep{Beaty2023}.
We investigate associative creativity in the naming and description of abstract visual figures, similar to prior work with stimuli such as tangrams \citep{Clark1986,ji2022abstract,Fasquel2022}.

\paragraph{Creativity in LLMs}

A lot of recent work investigates whether generative AI systems such as LLMs show signs of human-like creativity. 
   
Existing studies commonly assess creativity through creative-writing tasks \citep{Orwig2024, Ismayilzada2025, Li2025, BellemarePepin2026}
or through established metrics such as the \textit{Alternative Uses Task} (AUT, \citealt{Stevenson2022, Koivisto2023, Nath2025}), the \textit{Divergent Associations Task} (DAT, \citealt{Chen2023, BellemarePepin2026})
or combinations of the two \citep{Hubert2024, Haase2025}.
However, \citet{Nakajima2026} argue that these methods focus too heavily on novelty and are therefore insufficient for measuring creativity as a whole.
Other works propose novel benchmarks \citep{Dinu2024b}, some of which explicitly focus on associative creativity \citep{Qiu2025, Wadhwa2026}.
The evaluation results vary considerably: 
Some studies find that LLMs match or exceed average human performance
 \citep{Koivisto2023, Chen2023, Hubert2024, Orwig2024}, 
    
whereas others find that LLMs still do not reach human levels of creativity on more complex tasks \citep{Ismayilzada2025, BellemarePepin2026} 
and show qualitative differences in the responses \citep{Dinu2024,Nath2025} and did not increase significantly as a result of recent technical advancements \citep{Haase2025}.

\paragraph{Creativity in Vision-Language Settings}
\Sina{maybe focus on descriptions of abstract shapes in this part?}
Works such as KTH Tangrams \citep{Shore2018}, KiloGram \citep{ji2022abstract}, and SceneGram \citep{junker-zarriess-2025-scenegram} use abstract shapes such as tangrams to test multimodal generation systems on their ability to produce adequate descriptions. \Sina{we should also mention the color grid data set here, including our INLG paper... also, all of the following works seem less relevant to this study than the reference datasets mentioned in the previous sentence... however, we discuss the other ones more, than the reference datasets}
\citet{Yuan2025} identify a trade-off between faithfulness and creativity in MLLMs and propose a training-free approach to control associative behavior. 
Several studies introduce benchmarks for MLLM creativity \citet{Huang2025,Huang2025b,Fang2025,Liu2025}.

\paragraph{Research Gap}

None of the existing datasets explicitly account for variation in the difficulty of forming associative connections, for example due to greater shape complexity, lower recognizability, or lower degree of objecthood. 
Moreover, when abstraction abilities are evaluated, it is typically assumed that both models and humans must rely on abstraction-based strategies. In these studies, both annotators and models are typically instructed to name the figures, which largely restricts descriptions to associative language based on perceived similarities between the stimuli and familiar objects. In practice, however, they can often adopt qualitatively different strategies, such as bypassing holistic interpretations of a stimulus (e.g., names or descriptions of an entity it appears to depict) and instead describing more low-level properties of the stimulus. Thus, the present study uses more general instructions, and stimuli designed in a way that they can be described not only associatively, but also in more literal, non-associative ways.

\section{Methodology}
\label{sec:methodology}
This section introduces the two datasets, their collection and annotation, and the evaluation methods used in our analysis.

\subsection{Datasets}
\label{sec:datasets}

We present a new English naming dataset \textbf{PixCC} including visual stimuli that gradually increase in difficulty. We further extend the existing German naming dataset \textbf{SilCC} \citep{anonymous_osf} with MLLM-generated responses.

\begin{figure}[H]
    \centering

    \begin{subfigure}[t]{0.17\linewidth}
        \centering
        \includegraphics[width=\linewidth]{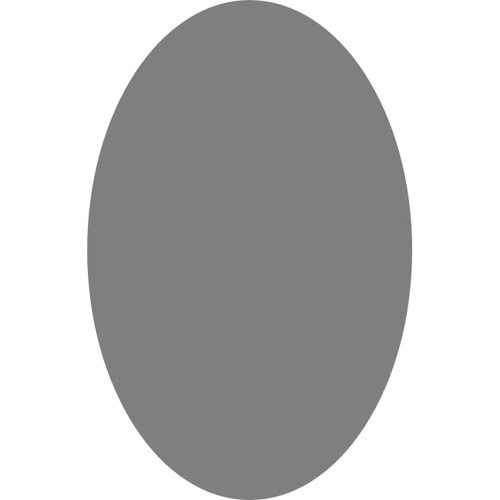}
        \caption{A}
    \end{subfigure}
    \hfill
    \begin{subfigure}[t]{0.17\linewidth}
        \centering
        \includegraphics[width=\linewidth]{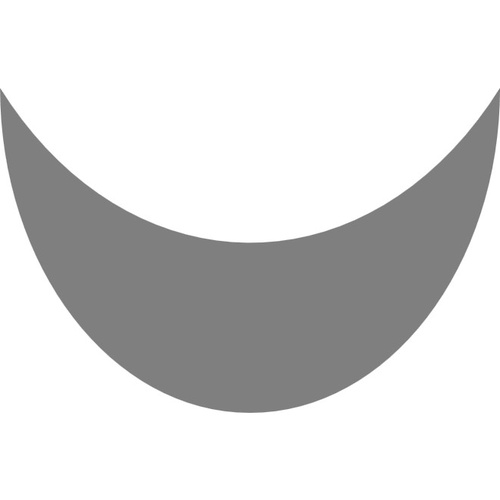}
        \caption{B}
    \end{subfigure}
    \hfill
    \begin{subfigure}[t]{0.17\linewidth}
        \centering
        \includegraphics[width=\linewidth]{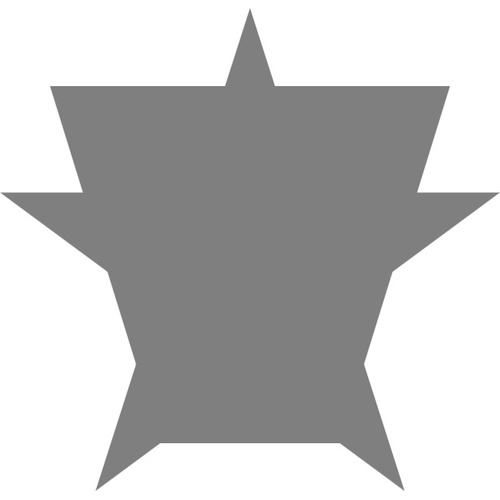}
        \caption{C}
    \end{subfigure}
    \hfill
    \begin{subfigure}[t]{0.17\linewidth}
        \centering
        \includegraphics[width=\linewidth]{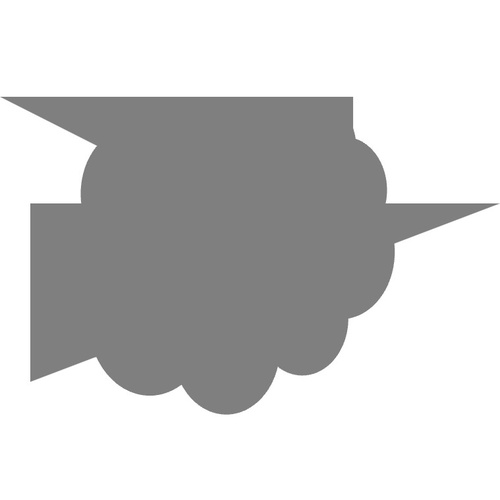}
        \caption{D}
    \end{subfigure}

    \vspace{0.5em}

    \begin{minipage}[t]{0.95\columnwidth}
    \footnotesize
    \textsf{\textbf{Human:}} 
        \textit{(a) Oval (oval), (b) Mond (moon), (c) Maske (mask), (d) Handshake (handshake)} \\
    \textsf{\textbf{LLaVA-34B:}} 
        \textit{(a) Kreis (circle), (b) Mondphase (moon phase), (c) Stern (star), (d) Lampenschirm (lampshade)} \\
    \textsf{\textbf{Qwen-72B:}}
        \textit{(a) Oval (oval), (b) Mond (moon), (c) Stern (star), (d) Sprechblase (speech bubble)} \\
    \textsf{\textbf{Qwen-235B:}}
        \textit{(a) Oval (oval), (b) Mondsichel (crescent moon), (c) Achtzackenstern (eight-pointed star
), (d) Sprechblase (speech bulb)} \\
    \textsf{\textbf{InternVL-38B:}}
        \textit{(a) Kreis (circle), (b) Halbmond (halfmoon), (c) Stern (star), (d) Wolke (cloud)}
    \end{minipage}
    \caption{Visual stimuli for each difficulty level from SilCC and examples for possible labels given by a source for each stimulus. Difficulty increases from A to D. 
    }
    \label{fig:SilCC_example}
\end{figure}

\paragraph{Silhouettes for Creativity in Communication (SilCC)}
The original SilCC collected 6,400 one-word references\footnote{The dataset also includes measures of Shannon entropy, fastText cosine similarity, Simpson’s diversity index, interpersonal agreement, and visual complexity, computed by the original authors. These measures are available in the associated OSF repository: \cite{anonymous_osf}.} by 160 L1 German speakers recruited on Prolific.
 The original item set includes 80 mono-chromatic, 2-dimensional visual stimuli that do not reliably correspond to a single lexical label designed to enable multiple plausible interpretations and encourage creative references \citep[]{anonymous_osf}.
 The stimuli comprise shapes at four levels of assumed difficulty: (A) familiar basic shapes, e.g., \textit{oval}; (B) familiar complex shapes, e.g., \textit{horizontal crescent}; (C) binary shape combinations, e.g., \textit{trapez}~$\cup$~\textit{star}; (D) unfamiliar shapes, e.g., \textit{cloud~$\cup$~right triangle~$\cup$~half frame}; see Figure~\ref{fig:SilCC_example} for one example from each level of difficulty.
Items were presented sequentially with no repetition in pseudo-randomized order. 

The participants' task was to name each object using a single word that describes the object as precisely as possible and were not explicitly instructed to be creative. 
The resulting study data was standardized in terms of capitalization and manually corrected for typing mistakes.

We extend SilCC by collecting image naming data from MLLMs, collecting human annotations of both human and model responses, and compute additional measures described in Section \ref{sec:annotation}.

\paragraph{Pixels for Creativity in Communication (PixCC)}

\begin{figure}
    \begin{center}
        \includegraphics[width=.17\columnwidth]{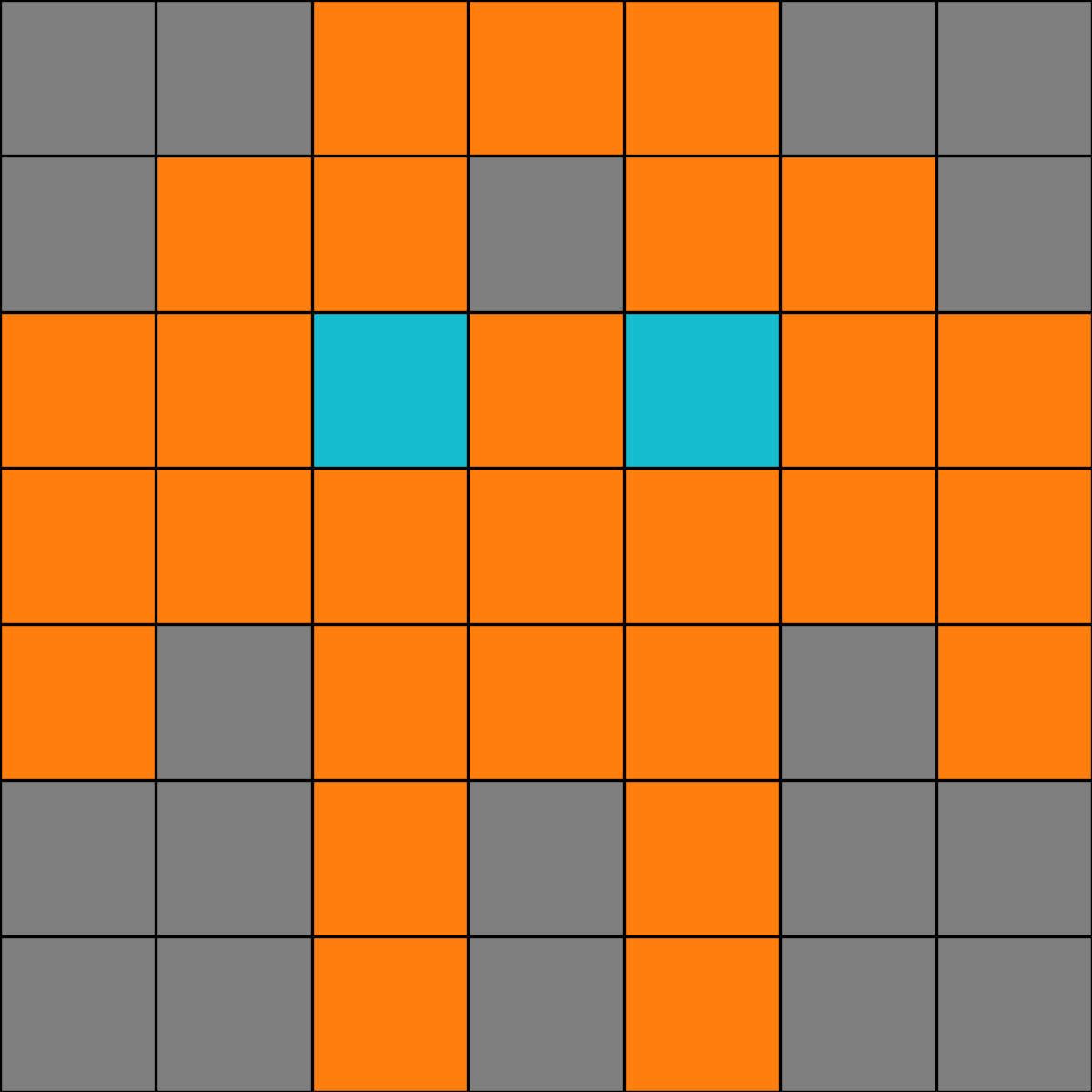}
        \includegraphics[width=.17\columnwidth]{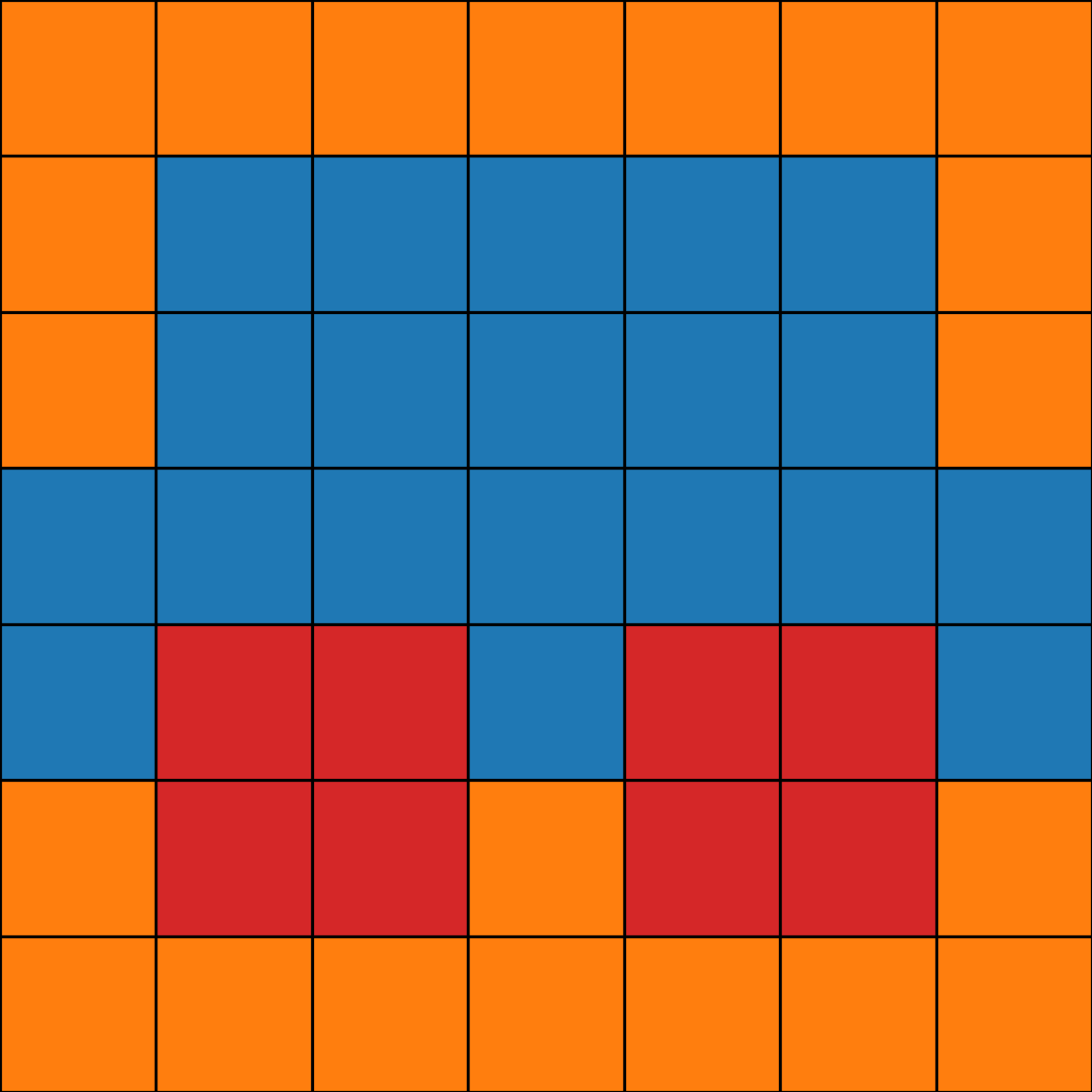}
        \includegraphics[width=.17\columnwidth]{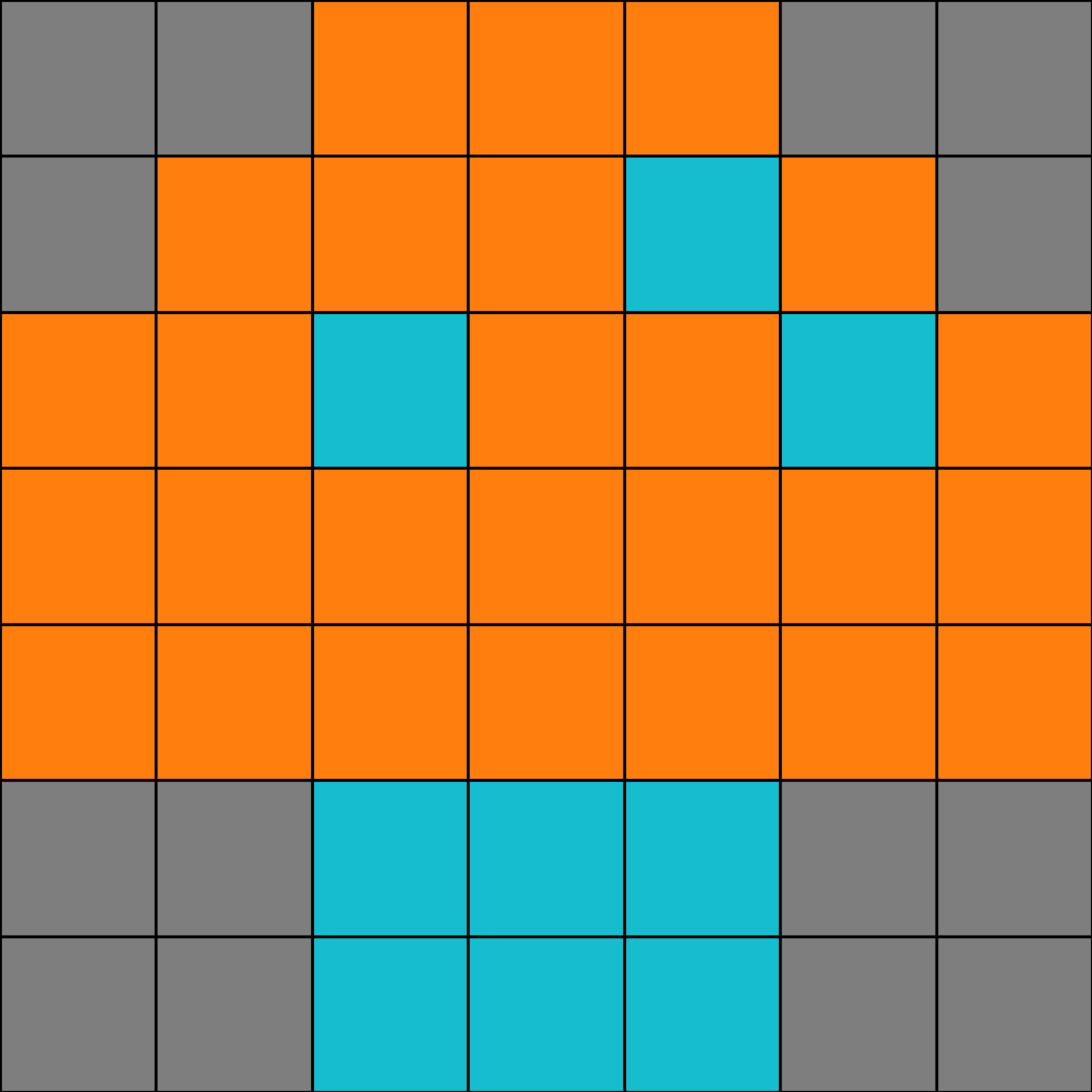}
        \includegraphics[width=.17\columnwidth]{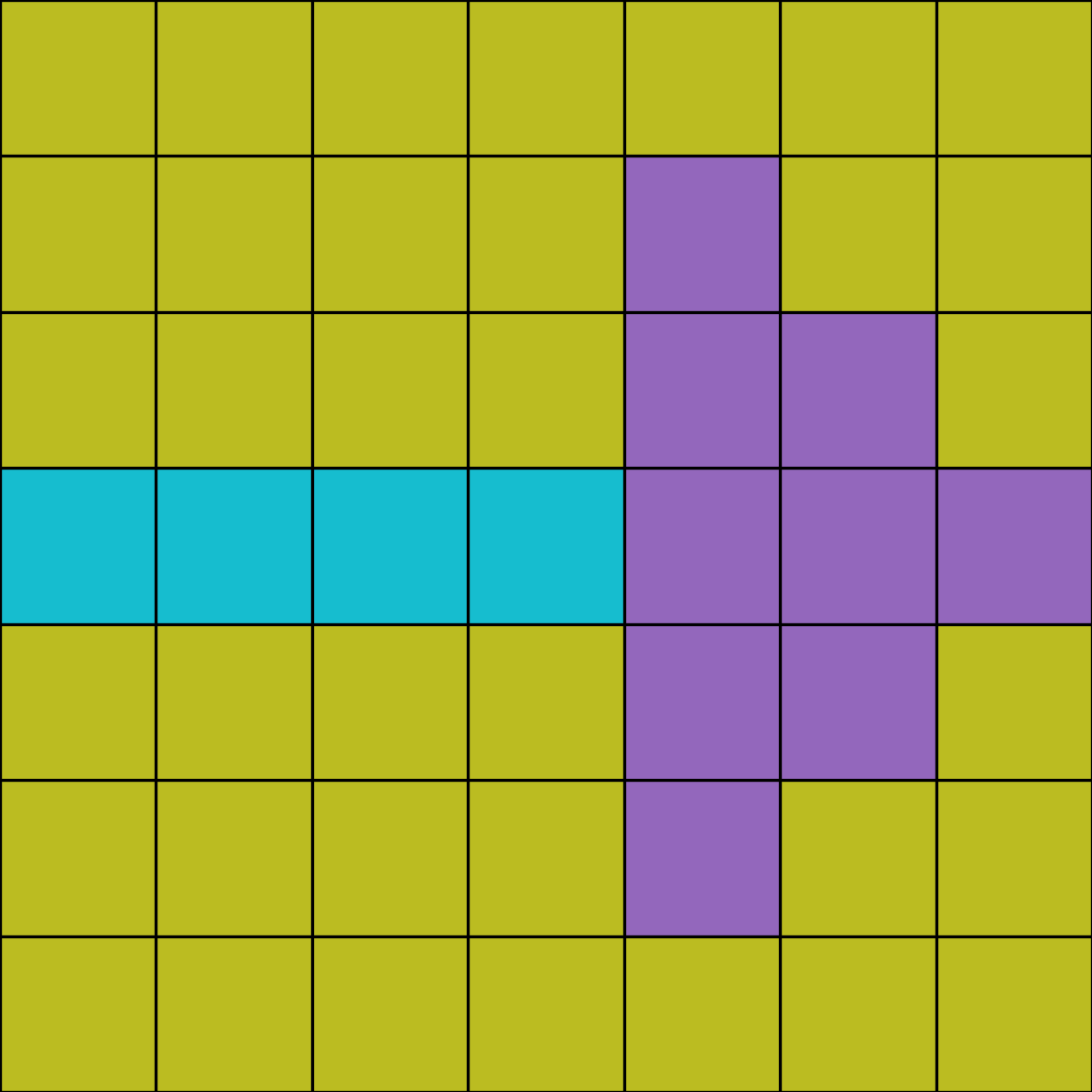}
    \end{center}
    \caption{Four types of stimuli presented in PixCC: creatures, objects, nature, and symbols. All stimuli here are unique clean items with no noise applied. Each stimulus was assigned one of three color palettes.} 
\label{fig:stimulipix}
\end{figure}

We created PixCC using a stimulus set consisting of 20 unique items, each represented as a $7 \times 7$ grid that depicts a pixelated shape drawn from one of four categories: creatures (e.g., fish, alien), objects (e.g., key, car), nature (e.g., tree, mushroom), and symbols (e.g., arrow, cross). We randomly sampled three color palettes of three colors each from the original ‘Tableau–10’ palette.
Each item was assigned one of the palettes, with one color serving as the background and two as foreground/object colors. We randomly sampled the color palettes to control for color diagnosticity and ensure that object identification cannot rely on color alone.  For instance, using the color yellow for a crescent like object might evoke a banana association, whereas a red round like object might evoke tomato. See Figure~\ref{fig:stimulipix} for one example of each shape category.

\begin{figure}
    \centering
    \includegraphics[width=0.5\textwidth]{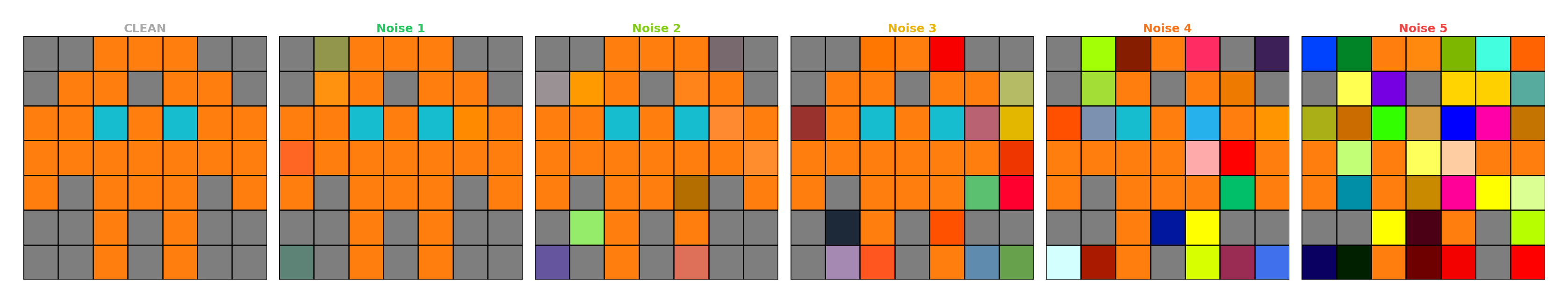}
    \caption{Gradual distortion of stimuli procedure presented in PixCC. The clean condition (far left) represents the original undistorted item. Noise increases progressively from left to right, with noise levels of 10\%, 20\%, 30\%, 40\%, and 70\% applied to the stimulus.}
    \label{fig:distortion}
\end{figure}

Naming difficulty is introduced through sparse Gaussian noise (percentage of pixels distorted) added both to the foreground and background of each image at five increasing levels: $10\%$, $20\%$, $30\%$, $40\%$, and $70\%$ (see Figure \ref{fig:distortion} for examples). At $70\%$ noise, images resemble randomly colored grids with no recognizable structure similar to the $3 \times 3$ color-grid stimuli used by \citet{mcdowell-goodman-2019-learning}. 
Combined with the original clean version, this yields six distortion levels per item, resulting in a total dataset of 120 stimuli (20 items $\times$ 6 noise levels). 

We recruited 67 L1 English speakers via Prolific, excluding colorblind individuals. 
The data was collected using oTree \citep{CHEN201688}. 
Each participant was assigned to one of six stimulus lists each containing 20 items. The lists contained one noise level per item, such that across all six lists, every item appeared at every noise level exactly once. Participants viewed the 20 items from their assigned lists and were instructed to give a name or a short description of what they saw by typing free text.  Additionally, participants completed a final naming task for one individually assigned ‘clean’ image with no noise added, providing 10 different names for this item. They were not explicitly instructed to be creative.
We collected a total of 1,340 descriptions and downsampled them to exactly 10 descriptions per item, prioritizing participants with  overall shorter descriptions in terms of number of words, resulting in a final dataset of 1,200 descriptions across 120 images and 60 participants. As for the final name task we got at least one list of names for each clean item.

\subsection{Models}
\label{sec:models}

We evaluate a set of open-weight MLLMs that vary in architecture and parameter scale, including Alibaba Cloud's Qwen3-VL-235B \citep{qwen3vltechnicalreport} and Qwen2.5-VL-72B \citep{qwen2.5vl} models, as well as LLaVA-v1.6-34B \citep{liu2024llavanext,liu2023improved} and InternVL3.5-38B \citep{InternVL}. All models are prompted with the same temperature (0.7) and receive the same task instructions as the human participants (see Appendix~\ref{app:prompts} for the complete prompts).
For each image, every model generated the same number of descriptions as collected in the corresponding human datasets. For SilCC each model produced 80 descriptions per image, resulting in 6,400 German descriptions per model, and for PixCC, each model produced 10 descriptions per image, resulting in 1,200 English descriptions per model. The presented approach is consistent with established practices for example in studies using the AUT \cite[cf.][]{GILHOOLY2024100071}. Similar procedures can be also found in computational creativity research \cite[e.g.][]{Nath2025, junker-zarriess-2025-scenegram, BellemarePepin2026}.

\subsection{Annotation}
\label{sec:annotation}

We collected human annotations for all descriptions from all sources (humans and models). The data were split into subsets, each of which was annotated by a single annotator familiar with the task. To assess inter-annotator agreement, we conducted a separate annotation study on a shared subset for both datasets involving all annotators. Further details are provided in Appendix~\ref{app:ann}.

For each description, annotators provided the following judgments:

\paragraph{Reflects} 
This annotation category captures whether the description accurately reflects the content of the image (label \textit{yes/no}). Annotators were further instructed to use the label \textit{okay} when they were genuinely uncertain if the description reflects the image. For SilCC, we also asked annotators to use the label \textit{okay} when the compositional nature of the stimulus was not fully reflected in the name, e.g., responses referring only to a subcomponent of the target stimulus' shape, e.g., “circle”, when the stimulus was of the type binary shape combination or unfamiliar shape. 

\paragraph{Semantic Head}

The semantic head represents a noun or compound that most accurately captures the primary concept conveyed by the description. This annotation was collected only for PixCC in order to make it comparable with SilCC. SilCC consists exclusively of one-word responses, including compound words that can become considerably long in German, semantic head annotation, hence, is largely redundant, as the lexical item itself already contains the semantic core of the utterance.  
    
\paragraph{Associativeness} 

This annotation category captures whether a description uses associative rather than literal descriptive language. We annotated this dimension for PixCC only. For example, the leftmost item in Figure~\ref{fig:stimulipix} can be described as a robot (high-level association) as well as by simply listing the colors in the grid.
In contrast, SilCC stimuli contain more or less recognizable object silhouettes, so one can naturally describe the perceived shape or object itself rather than low-level visual properties in one word, making an associativeness annotation less informative and less comparable to PixCC.

\paragraph{Novelty}

Novelty indicates how novel a given description is. In SilCC, due to the task specific constraint of only allowing for one-word responses,  we operationalize novelty straightforwardly at the lexical level. More precisely, we test if responses occur in the ``DWDS'' Dictionary of the German language \citep{dwds} or the Cambridge Dictionary \cite{cambridge_dictionary}. If the exact word was not found in both dictionaries, it was classified as \annolabel{novel}. 

In PixCC, we constructed a set of conventional names for each clean item. These names were collected from the human responses to the final naming task (see \ref{sec:datasets}). In addition to the human responses, we included the semantic head labels generated by each model for the clean version of each image in the conventional naming set. These descriptions serve as the baseline labels that both humans and models are expected to produce. We then checked for novelty by looking whether the semantic head for an image variant overlapped with the conventional names associated with that image. A semantic head was labeled as \annolabel{non-novel} if any lexical overlap was present, and as \annolabel{novel} otherwise. Before comparison, all text was lowercased and tokenised into individual words.

To account for repeated responses from the same source, we calculated a weighted novelty score by dividing it by the number of exact duplicate occurrences of the produced novel word within that source for the specific stimulus.

\subsection{Evaluation}
\label{sec:evaluation}
We evaluate both dataset jointly with respect to effectiveness and originality. We introduce the measures used and present the statistical models applied for analyses. By analyzing the datasets together, we aim to assess generalizability of the observed patterns across experimental settings.

\paragraph{Statistical modelling}

All models included \variable{source}, \variable{difficulty\_level} or \variable{noise\_level}, and their interaction as fixed effects. For models fitted to individual descriptions, we included participant random effects for human responses only, using \variable{is\_human} (1 for humans, 0 for models). Participant-specific slopes for difficulty or noise were modeled using \variable{human\_diff}, which equal the difficulty or noise level (rescaled from 0 to 1) for human responses and 0 for model responses. Random-effects structures initially included item and participant intercepts and by-participant slopes, and were simplified until convergence without singularity or correlations exceeding $|r| = .95$. Final structures are reported in table captions. We used sum coding for \variable{source}, polynomial contrasts for difficulty/noise, Type III Wald $\chi^2$ tests, and Tukey-adjusted pairwise comparisons.

\paragraph{Effectiveness}
We consider a description as effective if it accurately reflects the image content as labeled by the annotators. We analyzed \variable{reflectiveness} using the three ordered labels with cumulative link mixed-effects models (CLMMs), fitted to individual descriptions.

\paragraph{Originality}

We operationalized originality using two measures of semantic diversity, namely Shannon entropy \citep{Shannon1949} and semantic similarity, as well as tests for novelty and associativeness.
We used logistic mixed-effects models for associativeness at the description level and linear mixed-effects models for novelty and entropy at the item and source level, which include random intercepts for items only. The annotation of novelty and associativeness is described in Section~\ref{sec:annotation}.

To calculate response entropy, we calculated Shannon entropy across all answers/semantic heads for each item per source. In this context, higher entropy scores indicated a greater diversity of responses.

We calculate semantic similarity to assess the naming variability of a particular source with respect to all items at a given difficulty level. For this, we computed the cosine similarity between sentence embeddings generated using the all-MiniLM-L6-v2 model. 
 
For models, the semantic similarity was computed across all responses within a given difficulty level, pooling all responses from that sources at that difficulty. For human participants, we first calculated the mean pairwise cosine similarity within each participant's individual responses at a given difficulty level, then averaged these values across participants. Because this measure is computed per \variable{diffuclty\_level} and \variable{source} level, there are too few observations per source to support formal statistical modelling. We therefore treat semantic similarity as an exploratory measure and report descriptive trends only, without drawing inferential conclusions.

\section{Results}
This Section presents our results and analyses on effectiveness and originality of descriptions in the two datatsets.
Models were fitted as
described in Section~\ref{sec:evaluation}. The final random-effects structure of each model is given in the caption of the corresponding table. All referenced tables with
detailed statistics can be found in Appendix~\ref{app:add_results}.

\subsection{SilCC}
\label{sec:results:SilCC}

\paragraph{Effectiveness} 

\begin{figure*}
    \centering
    \includegraphics[width=.99\linewidth]{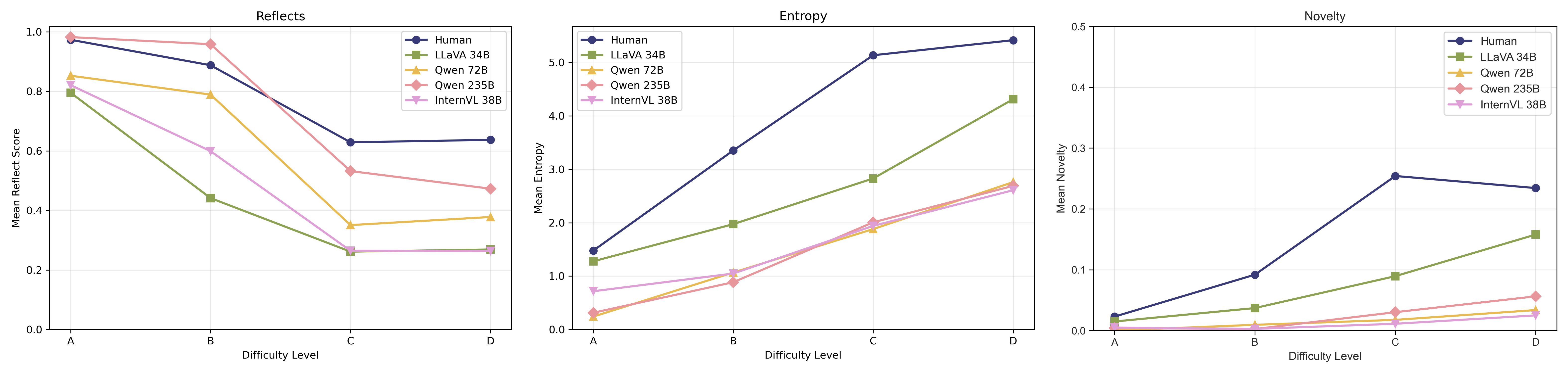}
    \caption{Result plots for reflects, entropy, and novelty in SilCC by difficulty level.}
    \label{fig:c02_combined}
\end{figure*}

 We find significant main effects on effectiveness for \variable{source} ($p<.001$), \variable{difficulty\_level} ($p<.001$) and their interaction ($p<.001$), see Tables~\ref{tab:clmm_successfulness}, ~\ref{tab:pairs_effectiveness} and~\ref{tab:anova_mr_g_clmm} for model results. As shown in Figure~\ref{fig:c02_combined}-left, the score of whether a one-word reference reflected the stimuli or not drops with rising difficulty level across all sources. A post-hoc analysis of estimated marginal means was conducted to compare sources within each difficulty level. This analysis reveals that Qwen-235B received higher scores at less difficult levels A and B. However, with rising difficulty level, humans outperform all models with significantly better reflectiveness scores.

\paragraph{Originality}

For novelty, we find significant main effects of \variable{source} ($p<.001$) and \variable{difficulty\_level} ($p<.001$), as well as their interaction ($p<.001$), see Tables~\ref{tab:lmm_novelty},~\ref{tab:pairs_by_difficulty_novelty} and~\ref{tab:anova_novelty} for model results.  
While human responses become significantly more novel with rising difficulty, models show weaker slopes, indicating that the effect of \variable{source} is highly dependent on \variable{difficulty\_level} (see Figure~\ref{fig:c02_combined}-right). Interestingly, the smallest model, LLaVA-34B, stands out, showing a steeper slope than
 the other models.

In terms of semantic similarity, we find a clear overall trend across all difficulty levels. The average human produces more semantically diverse responses than the models. However, human responses became slightly less diverse as item difficulty increased, though the overall similarity is constantly low across all levels. In contrast, the models show the opposite pattern, producing relatively less diverse responses for low-difficulty items but increasing in diversity with higher difficulty levels. Notably, each source produces a distinct pattern of semantic similarity across difficulty levels, indicating that there are source-specific response patterns rather than a uniform effect of difficulty. See Appendix \ref{app:add_results} (Figure \ref{fig:c02_semantic_sim}) a visualization of these trends.

When looking at entropy, we find significant main effects on \variable{source} ($p<.001$) and \variable{difficulty\_level} ($p<.001$) and their interaction ($p<.001$), see Tables~\ref{tab:lmm_entropy}, \ref{tab:pairs_by_difficulty}, and~\ref{tab:anova_entropy} for model results. While human responses become significantly more diverse with rising difficulty, models show weaker slopes, indicating that the effect of \variable{source} is highly dependent on \variable{difficulty\_level} (see Figure~\ref{fig:c02_combined}-center). Similar to the pattern  observed for novelty, LLaVA-34B again stands out, producing a higher variance in words and outperforming the other models at highest difficulty level ($p<.001$ for all 3 models).

Further analysis on originality measures conditioned on effectivenes confirms these findings (Appendix \ref{app:creativity}).

\subsection{PixCC}
\label{sec:results:pixcc}

\begin{figure*}
    \centering
    \includegraphics[width=.99\textwidth]{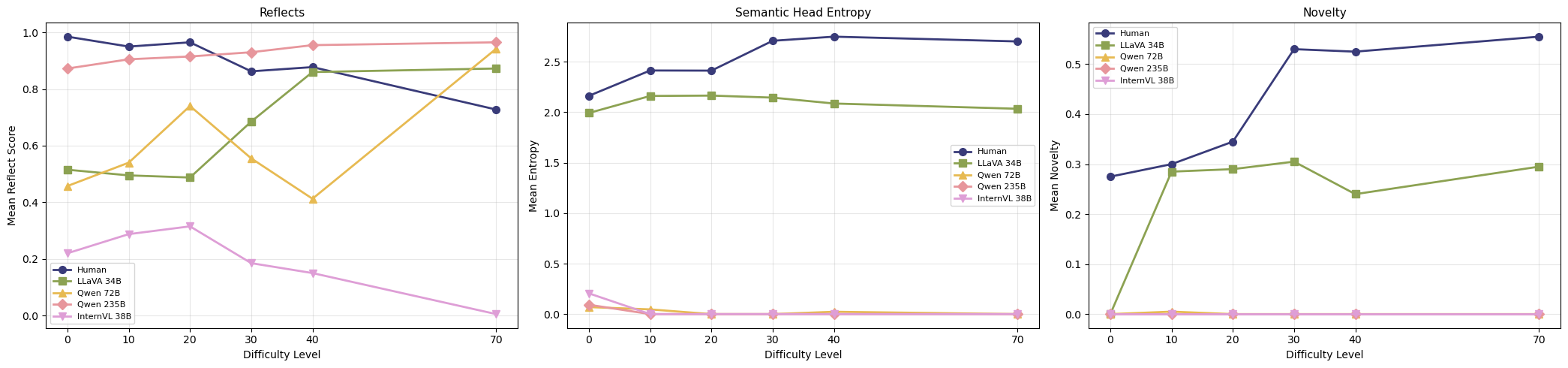}
    \caption{Result plots for reflects and entropy (on the full data) and novelty in PixCC by difficulty level.}
    \label{fig:results_pixcc}
\end{figure*}

\paragraph{Effectiveness}
We find significant effects of \variable{source} ($p<.001$), \variable{noise\_level} ($p=.032$), and their interaction ($p<.001$), see Tables~\ref{tab:anova_eff_pixcc} and~\ref{tab:pairs_eff_pixcc}. The significant interaction indicates that the effect of
noise on reflectiveness differs across sources.
As shown in Figure~\ref{fig:results_pixcc}-left, in the easiest condition, when the images are clean, humans get the highest reflectiveness score and significantly outperform
all models (all $p<.001$), followed by the large Qwen-235B model. Human performance drops with increasing difficulty, consistent with the expected effect of image degradation on the ability to produce grounded descriptions. However, this trend is largely reversed for the models (except for InternVL-38B). For instance, LLaVA-34B and the smaller Qwen-72B perform poorly in the easier conditions and remain
significantly below humans up to 30\% noise for LLaVA-34B and to 40\% for Qwen-72B, but show higher reflectiveness scores at the highest noise level. Qwen-235B does not differ significantly from humans at
10-30\% noise, but significantly outperforms them at 40\% ($p=.001$) and 70\% ($p<.001$). At 40\% noise, LLaVA-34B and humans do not differ significantly ($p=1.000$), while at 70\% noise, LLaVA-34B and Qwen-72B also significantly
outperformed humans (both $p<.001$). InternVL-38B performed poorly overall and was rated significantly lower than humans at every noise level (all $p<.001$), as most
of its answers were labeled non-reflective, primarily due to counting and positioning errors within the grid, such as frequently describing a $5\times5$ or $9\times9$ grid. Figure~\ref{fig:PixRef} illustrates these differences.

\paragraph{Originality}

For associativness, we also find significant main effects for \variable{source}, \variable{noise\_level} and their interaction (all $p<.001$). As shown in Figure~\ref{fig:succ_asso_full}, humans produce associative descriptions under the clean condition. However, as the items degrade due to noise, their descriptions become significantly less associative. In this dataset, associativity is optional, and humans often rely on alternative strategies when the items are at higher noise levels. 
In particular, they often incorporate low-level properties into their descriptions of degraded items, as in ``a complete mixture of lots of different colors''.

In contrast, LLaVA-34B consistently produces associative descriptions (across all noise levels and significantly more often than humans at 40\% and 70\%), although we qualitatively observe that these associations are highly generic and repetitive across items. For example, the model frequently used the association ``mosaic''. The other models produce significantly fewer associative descriptions overall (except at 70\%, where Qwen-72B and InternVL-38B do not differ from humans). The smaller Qwen-72B and InternVL-38B models use some associative descriptions under clean conditions, but their use decreased as noise increased, except for the 70\% noise condition, where associativity increases again due to repeated ``mosaic'' descriptions. The large Qwen-235B almost never use associative descriptions, except for few cases in the clean condition. A summary of the model results can be found in Appendix~\ref{app:assoc}.

For Novelty, we find significant main effects for \variable{source} ($p<.001$) and \variable{noise\_level} ($p=.002$) as well as significant interaction ($p<.001$). As shown in Figure \ref{fig:results_pixcc} (right), humans produced more novel semantic heads with increasing difficulty, while LLaVA-34B shows relatively little variation. 
Humans produce significantly more novel associations than LLaVA-34B at 30\%, 40\%, and 70\% noise (all $p<.001$), but not at 10\% or 20\%.
For novelty the linear mixed-effect model was restricted to humans and LLaVA-34B  at noise levels 10–70\% as the other models produce (almost) no novel semantic heads, and model responses to clean images are non-novel by construction, since their own semantic heads are part of the conventional naming set.
 
A summary of the model results can be found in Appendix~\ref{app:nov}. 

In terms of semantic similarity, we also find a clear overall trend across all difficulty levels. The average human produces considerably more semantically diverse responses than the models. Here, human responses also became slightly less diverse as item difficulty increased, though the overall similarity is constantly low across all levels. Models show similar tendencies with increasing difficulty levels. A visualization of these trends can be found in Appendix~\ref{app:add_results} in Figure~\ref{fig:B02_semantic_sim}.

For the semantic head entropy, we find significant effects for \variable{source} ($p<.001$) and a significant interaction with \variable{noise\_level} ($p=.003$) but no significant main effect of \variable{noise\_level}, see a summary of the results in Appendix \ref{app:ent}. As shown in Figure~\ref{fig:results_pixcc}-center, humans show the most diverse semantic heads overall, with entropy increasing significantly with noise level. LLaVA-34B also produces
diverse semantic heads, comparable to humans at 0--20\% noise but significantly less diverse at 30--70\% (all $p<.001$). However, in a model restricted to these two
sources, the interaction was not significant. The remaining models show near-zero entropy across all noise levels. 
Further analysis on originality measures conditioned on effectiveness confirmed these findings (see \ref{app:creativity})

\section{Discussion}
\label{sec:discussion}

In this work, we investigate the capabilities of humans and MLLMs in creative naming tasks under gradually increasing stimulus difficulty. 
In SilCC, associativity is inherently required because the requirement to describe the figures in one word demands holistic conceptualizations. On the other hand in PixCC associativity is optional, since the images can be described not only in terms of their resemblance to familiar objects, but also on the basis of more granular low-level properties.

Our findings show that MLLMs generally fall short of human performance, particularly with respect to \textit{originality}. Across both datasets, humans produced more novel descriptions, especially at higher difficulty levels. Most models, except for LLaVA-34B, generated very few novel descriptions, resulting in lower overall originality compared to humans. This distinction is further supported by our semantic diversity analysis.

Regarding \textit{effectiveness}, MLLMs often fail to achieve human performance in SilCC. The task-specific constraint of naming all items using only a single word requires associative reasoning throughout the whole task. While items at lower difficulty levels can often be named relatively straightforwardly, more difficult stimuli require stronger associative and abstract reasoning capabilities.
However, this pattern does not generally hold for PixCC: in this dataset, humans perform best in the clean condition, and their performance remains competitive, although it declines, as task difficulty increases. Interestingly, most models show the opposite trend: although they initially perform worse than humans, their performance improves as task difficulty increases. 
This might be related to the observation that the descriptions generated by all MLLMs except for LLaVA-34B are primarily based on low-level visual features instead of high-level associations, which might be a more robust strategy in high-noise settings. 
Importantly, humans are remarkably flexible in their description strategies, i.e., they switch from associative to low-level descriptions as noise increases, 
whereas the associativeness patterns found in MLLMs appear less systematic. 
This suggests that creative problem solving abilities beyond the lexical-semantic level are required for this task, the investigation of which remains an important topic for future research.

Finally, model size also appears to play a role in task performance, especially in SilCC, where larger models outperform smaller ones, with the largest model evaluated in this study (Qwen-235B) achieving the highest overall performance. However, in PixCC, this trend is only apparent for the largest model, while no clear relationship between model size and performance can be observed among the remaining models. Interestingly, however, the smallest model stands out with respect to novelty and semantic diversity, a finding similar to \citet[p.~2647]{Nakajima2026}, who observed that ``simpler families tend toward higher novelty with reduced appropriateness''.
Still, humans remain clearly distinct from all evaluated sources in their ability to generate novel and diverse descriptions. 

\section{Conclusion}
\label{sec:conclusion}

In this paper, we presented two datasets designed to study human creativity in naming tasks under increasing difficulty: an extended SilCC \citep{anonymous_osf}, and the newly collected dataset PixCC. Using these datasets, we examined whether MLLMs can generate creative associations similar to those produced by humans. Our findings show important differences between humans and MLLMs in naming tasks. Humans consistently produce more original descriptions than models at higher difficulty levels and appear to be unique in balancing both effectiveness and originality, combining successful associative naming with the generation of novel yet meaningful references. Our results suggest that current state-of-the-art MLLMs still struggle with associative conceptualization, as they produce less original and less diverse associations than humans and do not fully capture the flexible and creative language abilities of humans.

\section*{Limitations}
\label{sec:limitations}

A limitation of the present paper is that human participants were presented with multiple instances, which may have introduced training or familiarity effects over time. 

Additionally, our qualitative observation of the human responses in PixCC suggests that some of the descriptions were AI generated, as these responses were unusually long, detailed and repetitive. Therefore, we prioritized participants with overall shorter descriptions by word count. Furthermore, some participants did not complete the final naming task with the required 10 names resulting in an imbalance across items within the conventional names set.

On the model side, our data collection was conducted under the exact same generation parameter for each model. In particular, varying the temperature could affect the diversity and consistency of responses and should be systematically explored in future analysis.

We also observed that LLaVA-34B often produced a more than one interpretation for the target stimuli and at times shifted to unrelated topics (e.g., generally discussing what pixel art is), which made it difficult to identify a single semantic head for comparison. In addition, outputs often relied on generic associations such as shapes, mosaics, or retro video game aesthetics, resulting in largely associative descriptions labels.

A limitation of SilCC is that it does not allow us to fully disentangle perceptual error from associative failures. An annotation of 0 indicates that a label does not reflect the target stimulus, but this may result from incorrect shape recognition, placeholder or all porpouse words, or associations that are not grounded in the stimulus. However, our goal is to evaluate associative creativity in MLLMs end-to-end, where the ability to effectively interpret visual input is an inherent component of the task (creative cognition combines bottom-up perceptual processes and top-down control; see, e.g., \citet{Beaty2016CreativeCognition}). Therefore, failures in visual grounding can also affect the effectiveness of the generated descriptions.

At the time of conducting this study, we deliberately chose not to expose our dataset to commercial models and instead focus on local open-weight systems in order to ensure greater control and reproducibility. We nevertheless acknowledge that comparisons with frontier models could yield valuable additional insights and we consider such comparisons to be a promising direction for future work.

Furthermore, our evaluation compared individual human responses against multiple responses generated by the same models, creating an asymmetry in the evaluation. A more balanced setting would involve collecting multiple responses from the same human participants as well. However, here we follow the common practice in computational creativity research, particularly studies using Alternative Uses Tasks.

Finally, the paper relied on only three annotators per dataset, which limits the generalization of the annotations. Future work should therefore include a larger annotator pool in order to reduce the influence of individual biases. For PixCC, the shared annotation subset does not include samples from all tested models.

\section*{Acknowledgments}
We are grateful to the student assistants Svenja Ingmanns, Alice Schirrmeister, Larissa Koch, Florian Steig, Vicky Rupprecht for supporting the study with annotations. We would also like to thank Annett Jorschick for her assistance with the statistical analysis.\\
This research has been funded by the Deutsche Forschungsgemeinschaft (DFG, German Research Foundation) – CRC-1646, project number 512393437, project B02 and C02.

\bibliography{bibliography-emnlp-B02-C02}

\appendix

\section{CRediT Contributor Statement}
\insertcreditsstatement

\section{Risks and Ethical Considerations}
We do not believe that there are significant risks
associated with this work, as we analyze existing
models using data with limited scale without con-
tents that might be perceived as hurtful. No directly identifying information was collected in this study. No general ethics review was required. However, the human data collection in SilCC received ethics approval from the university ethics review board prior to data collection.

\section{Datasets}
\label{sec:appendix}
\subsection{SilCC}

SilCC contains 6,400 one-word responses given by 160 L1 German speakers which were recruited on Prolific (62 female, 98 male; mean age = 35.92 [SD = 10.57]), and paid 2 GBP for participation. The study was conducted following ethics guidelines, and informed consent was given from all participants before participation.

\subsection{PixCC}
\label{app:pixcc}

PixCC contains 1,340 descriptions collected from 67 participants recruited via prolific (the retained analysis sample was 1200 descriptions). 
The experiment was implemented in otree \citep{CHEN201688} using a custom web-based interface hosted on a university server via Nginx. To ensure data quality, the study included attention checks. All participants provided informed consent, including consent for their data to be used for research purposes. Participants were compensated in accordance with Prolific's recommended fair-pay guidelines.

\section{Prompts}
\label{app:prompts}

\subsection{SilCC}

\paragraph{SilCC Prompt}
\hfill\\
{\small\ttfamily\noindent
\begin{tabularx}{\linewidth}{X}
\toprule
    Im Folgenden wird Ihnen ein Bild von einem Objekt präsentiert (es wird Ihnen entweder bekannt vorkommen, oder weniger).\\
    Ihre Aufgabe ist es, das Objekt mit einem Wort zu benennen, das dieses Objekt möglichst genau beschreibt.\\
    Gib NUR das Wort zurück, ohne Erklärung oder zusätzlichem Text \footnote{For SilCC we further instructed the model to return only the target word, without any explanation or additional text, so that its output format would match the human responses}. Antworten Sie auf DEUTSCH\footnote{For InternVL-38B, LLaVA-34B and Qwen-235B we additionally instructed to answer in German only}.\\
\bottomrule
\end{tabularx}}

\paragraph{SilCC Prompt (translated)}
\hfill\\
{\small\ttfamily\noindent
\begin{tabularx}{\linewidth}{X}
\toprule
    Below you will see a picture of an object (it will either look familiar to you or not).\\
    Your task is to name the object with a single word that describes it as accurately as possible.\\
    ONLY provide the word, without any explanation or additional text. Please respond in GERMAN\footnotemark[4].\\
\bottomrule
\end{tabularx}}

\subsection{PixCC}
\paragraph{PixCC Prompt}
\hfill\\
{\small\ttfamily\noindent
\begin{tabularx}{\linewidth}{X}
\toprule
    Provide a name or a short description of what you see. The description should allow someone else to identify the image among similar images.\\
\bottomrule
\end{tabularx}}

\section{Annotation}
\label{app:ann}
All annotators were german speaking student assistants (5 females and one male).

\subsection{SilCC}

 Three independent annotators labeled the responses in our dataset. To reduce redundancy, we searched the full dataset of 32,000 responses for duplicated answers within each item and aggregated identical responses, resulting in 6,662 unique answers. From these, we randomly sampled a subset of 1,069 answers to assess interrater reliability. For the reliability analysis, we excluded all samples that received a rating of \textit{Okay=0.5} from at least one annotator, as this value indicated high annotator uncertainty. After exclusion, 684 samples remained for analysis. We then calculated Fleiss' $\kappa$ to assess interrater agreement, resulting in a value of $\kappa=0.68$ and indicating good reliability among annotators. Beyond the shared subset used for reliability estimation, Annotators 1 and 2 each annotated and additional subset of 1,300 samples, while Annotator 3 annotated 2,994 samples. For the analyses reported in this paper, we included the annotations from Annotator 3 for the samples that were part of the interrater reliability subset.

\subsection{PixCC}

Three annotators labeled PixCC for reflectiveness, associativeness, and semantic heads. 
All annotators labeled a shared subset of the dataset consisting of 144 descriptions in total, drawn from a mixture of human responses (90\%) and responses generated by the large Qwen-235B model (10\%).  Inter-annotator agreement for the reflective label was measured using raw percent agreement (89.3\%), as the near absence of ``No'' labels makes Fleiss' $\kappa$ an unreliable measure for this category. 
Agreement for semantic head annotations was high, with Fleiss' $\kappa = 0.62$. 
For associativeness, agreement on the binary yes/no labels was also high, with Fleiss' $\kappa = 0.86$.

\section{Additional Results}
\label{app:add_results}

\begin{figure}[]
    \centering
    \includegraphics[width=0.3\linewidth]{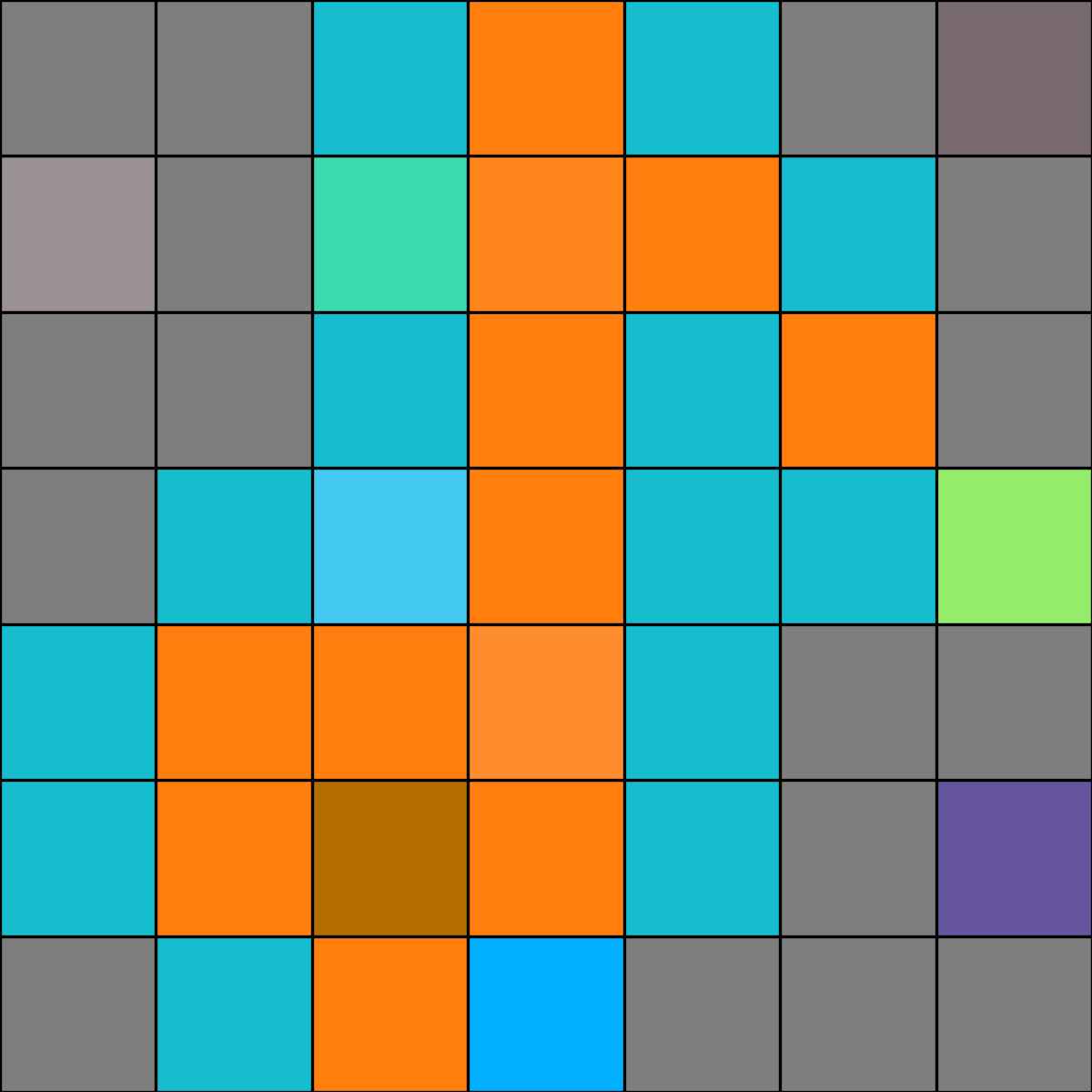}
        \begin{minipage}[t]{0.95\columnwidth}\footnotesize{
    \textsf{\textbf{Human:}} 
        \textit{“a giraffe on a partly cloudy day”} \\
    \textsf{\textbf{LLaVA-34B:}} 
        \textit{“This is an abstract digital art piece featuring a grid with various square tiles, each filled with different colors such as blue, orange, red, green, and purple. The arrangement of squares does not form any identifiable figures or objects; instead, it appears to be a random or deliberate pattern of colored squares against a dark background that could also be interpreted as additional gray squares forming the grid. The overall style is minimalist with a focus on color contrast and simplicity.”}\\
    \textsf{\textbf{Qwen-72B:}}
        \textit{“This image is a 7x7 grid with colored squares. The colors used are gray, cyan, orange, green, and purple. The pattern is irregular, with no obvious repeating sequence. Notable features include a single green square in the bottom right section and a single purple square in the bottom right corner. The rest of the squares are filled with gray, cyan, and orange in a seemingly random arrangement.”}\\
    \textsf{\textbf{Qwen-235B:}}
        \textit{“A 7x7 grid with a central cluster of orange and teal squares, surrounded by gray squares, featuring distinct brown, purple, and green squares in specific positions.”}\\
    \textsf{\textbf{InternVL-38B:}}
        \textit{“The image is a 9x9 grid with various colored squares, including shades of blue, orange, brown, green, and purple, arranged in a pattern. The colors are not uniformly distributed, creating a mosaic-like design.”}
    }\end{minipage}
    \caption{%
        PixCC example responses for an item (category symbol, item: musical note) with 20\% noise added.
    }
    \label{fig:PixRef}
\end{figure}

\subsection{Associativeness}
\label{app:assoc}
\begin{table}[H]
\scriptsize
\setlength{\tabcolsep}{1.5pt}
\centering
\resizebox{0.5\textwidth}{!}{%
\begin{tabular}{lrrrll}
\toprule
\textbf{Coefficients} & \textbf{Estimate} & \textbf{Std. Error} & \textbf{z} & \textbf{OR (95\% CI)} & \textbf{p} \\
\midrule
(Intercept) & -1.20656 & 0.14104 & -8.555 & 0.30 [0.23, 0.39] & \textbf{< .001} \\
Human & 2.29927 & 0.47783 & 4.812 & 9.97 [3.91, 25.43] & \textbf{< .001} \\
LLaVA 34B & 2.89487 & 0.14858 & 19.484 & 18.08 [13.51, 24.19] & \textbf{< .001} \\
Qwen 72B & -0.80519 & 0.15457 & -5.209 & 0.45 [0.33, 0.61] & \textbf{< .001} \\
Qwen 235B & -2.99458 & 0.24812 & -12.069 & 0.05 [0.03, 0.08] & \textbf{< .001} \\
noise\_level.L & -0.90766 & 0.17094 & -5.310 & 0.40 [0.29, 0.56] & \textbf{< .001} \\
noise\_level.Q & 1.09986 & 0.15949 & 6.896 & 3.00 [2.20, 4.11] & \textbf{< .001} \\
noise\_level.C & -0.14905 & 0.15241 & -0.978 & 0.86 [0.64, 1.16] & .328 \\
noise\_level\textasciicircum 4 & -0.02526 & 0.18706 & -0.135 & 0.98 [0.68, 1.41] & .893 \\
noise\_level\textasciicircum 5 & -0.20356 & 0.18827 & -1.081 & 0.82 [0.56, 1.18] & .280 \\
Human:noise\_level.L & -3.07024 & 0.47032 & -6.528 & 0.05 [0.02, 0.12] & \textbf{< .001} \\
LLaVA 34B:noise\_level.L & 0.53927 & 0.22620 & 2.384 & 1.71 [1.10, 2.67] & \textbf{.017} \\
Qwen 72B:noise\_level.L & 0.62627 & 0.22286 & 2.810 & 1.87 [1.21, 2.90] & \textbf{.005} \\
Qwen 235B:noise\_level.L & 0.23432 & 0.45067 & 0.520 & 1.26 [0.52, 3.06] & .603 \\
Human:noise\_level.Q & 0.07898 & 0.28106 & 0.281 & 1.08 [0.62, 1.88] & .779 \\
LLaVA 34B:noise\_level.Q & -1.05359 & 0.22005 & -4.788 & 0.35 [0.23, 0.54] & \textbf{< .001} \\
Qwen 72B:noise\_level.Q & 0.50923 & 0.23768 & 2.143 & 1.66 [1.04, 2.65] & \textbf{.032} \\
Qwen 235B:noise\_level.Q & 0.19152 & 0.50344 & 0.380 & 1.21 [0.45, 3.25] & .704 \\
Human:noise\_level.C & 0.81411 & 0.26036 & 3.127 & 2.26 [1.35, 3.76] & \textbf{.002} \\
LLaVA 34B:noise\_level.C & 0.09028 & 0.21455 & 0.421 & 1.09 [0.72, 1.67] & .674 \\
Qwen 72B:noise\_level.C & -0.21669 & 0.23818 & -0.910 & 0.81 [0.50, 1.28] & .363 \\
Qwen 235B:noise\_level.C & -0.71088 & 0.46288 & -1.536 & 0.49 [0.20, 1.22] & .125 \\
Human:noise\_level\textasciicircum 4 & 0.28286 & 0.27376 & 1.033 & 1.33 [0.78, 2.27] & .301 \\
LLaVA 34B:noise\_level\textasciicircum 4 & 0.05652 & 0.24364 & 0.232 & 1.06 [0.66, 1.71] & .817 \\
Qwen 72B:noise\_level\textasciicircum 4 & -0.41021 & 0.26940 & -1.523 & 0.66 [0.39, 1.13] & .128 \\
Qwen 235B:noise\_level\textasciicircum 4 & 0.01794 & 0.62714 & 0.029 & 1.02 [0.30, 3.48] & .977 \\
Human:noise\_level\textasciicircum 5 & -0.24709 & 0.27646 & -0.894 & 0.78 [0.45, 1.34] & .371 \\
LLaVA 34B:noise\_level\textasciicircum 5 & 0.46915 & 0.24130 & 1.944 & 1.60 [1.00, 2.57] & .052 \\
Qwen 72B:noise\_level\textasciicircum 5 & 0.36895 & 0.29114 & 1.267 & 1.45 [0.82, 2.56] & .205 \\
Qwen 235B:noise\_level\textasciicircum 5 & -0.71773 & 0.61919 & -1.159 & 0.49 [0.14, 1.64] & .246 \\
\bottomrule
\end{tabular}
}
\caption{Logistic mixed model for \textit{associativeness} in PixCC: $\variable{associative} \sim \variable{source} * \variable{noise\_level} + (0 + \variable{is\_human} + \variable{human\_diff} \mid \variable{participant}) + (1 \mid \variable{image})$.}
\label{tab:coef_assoc_pixcc}
\end{table}

\begin{table}[H]
\scriptsize
\setlength{\tabcolsep}{6pt}
\centering
\begin{tabular}{lrrl}
\toprule
\textbf{Effect} & $\boldsymbol{\chi^2}$ & \textbf{df} & \textbf{p} \\
\midrule
source & 1454.90 & 4 & \textbf{< .001} \\
noise\_level & 92.06 & 5 & \textbf{< .001} \\
source:noise\_level & 123.18 & 20 & \textbf{< .001} \\
\bottomrule
\end{tabular}
\caption{Type III Wald $\chi^2$ tests for \textit{associativeness} in PixCC. Logistic mixed model: $\variable{associative} \sim \variable{source} * \variable{noise\_level} + (0 + \variable{is\_human} + \variable{human\_diff} \mid \variable{participant}) + (1 \mid \variable{image})$}
\label{tab:anova_assoc_pixcc}
\end{table}

\begin{table}[H]
\scriptsize
\setlength{\tabcolsep}{1.5pt}
\centering
\resizebox{0.5\textwidth}{!}{%
\begin{tabular}{llrrl}
\toprule
\textbf{Noise} & \textbf{Contrast} & \textbf{Odds ratio} & \textbf{95\% CI} & \textbf{p} \\
\midrule
0\% & Human / LLaVA 34B & 5.29 & [0.508, 55.1] & .296 \\
 & Human / Qwen 72B & 90.7 & [8.99, 916] & \textbf{< .001} \\
 & Human / Qwen 235B & 524 & [47.4, 5800] & \textbf{< .001} \\
 & Human / InternVL 38B & 310 & [29.3, 3280] & \textbf{< .001} \\
 & LLaVA 34B / Qwen 72B & 17.1 & [8.36, 35.1] & \textbf{< .001} \\
 & LLaVA 34B / Qwen 235B & 99.0 & [37.4, 262] & \textbf{< .001} \\
 & LLaVA 34B / InternVL 38B & 58.6 & [24.9, 138] & \textbf{< .001} \\
 & Qwen 72B / Qwen 235B & 5.78 & [2.37, 14.1] & \textbf{< .001} \\
 & Qwen 72B / InternVL 38B & 3.42 & [1.60, 7.33] & \textbf{< .001} \\
 & Qwen 235B / InternVL 38B & 0.592 & [0.217, 1.62] & .612 \\
\midrule
10\% & Human / LLaVA 34B & 1.77 & [0.217, 14.5] & .946 \\
 & Human / Qwen 72B & 58.9 & [7.30, 476] & \textbf{< .001} \\
 & Human / Qwen 235B & 1160 & [70.5, 19100] & \textbf{< .001} \\
 & Human / InternVL 38B & 201 & [22.4, 1800] & \textbf{< .001} \\
 & LLaVA 34B / Qwen 72B & 33.2 & [15.4, 71.7] & \textbf{< .001} \\
 & LLaVA 34B / Qwen 235B & 653 & [86.9, 4910] & \textbf{< .001} \\
 & LLaVA 34B / InternVL 38B & 113 & [40.9, 314] & \textbf{< .001} \\
 & Qwen 72B / Qwen 235B & 19.7 & [2.65, 146] & \textbf{< .001} \\
 & Qwen 72B / InternVL 38B & 3.41 & [1.26, 9.21] & \textbf{.007} \\
 & Qwen 235B / InternVL 38B & 0.173 & [0.021, 1.43] & .157 \\
\midrule
20\% & Human / LLaVA 34B & 1.71 & [0.253, 11.5] & .941 \\
 & Human / Qwen 72B & 125 & [16.7, 938] & \textbf{< .001} \\
 & Human / Qwen 235B & 525 & [46.2, 5980] & \textbf{< .001} \\
 & Human / InternVL 38B & 170 & [21.5, 1340] & \textbf{< .001} \\
 & LLaVA 34B / Qwen 72B & 73.3 & [28.1, 191] & \textbf{< .001} \\
 & LLaVA 34B / Qwen 235B & 308 & [58.2, 1620] & \textbf{< .001} \\
 & LLaVA 34B / InternVL 38B & 99.3 & [34.4, 286] & \textbf{< .001} \\
 & Qwen 72B / Qwen 235B & 4.20 & [0.706, 24.9] & .181 \\
 & Qwen 72B / InternVL 38B & 1.36 & [0.394, 4.66] & .963 \\
 & Qwen 235B / InternVL 38B & 0.323 & [0.051, 2.03] & .448 \\
\midrule
30\% & Human / LLaVA 34B & 0.230 & [0.039, 1.35] & .156 \\
 & Human / Qwen 72B & 25.0 & [3.80, 164] & \textbf{< .001} \\
 & Human / Qwen 235B & 289 & [11.8, 7120] & \textbf{< .001} \\
 & Human / InternVL 38B & 34.9 & [4.97, 244] & \textbf{< .001} \\
 & LLaVA 34B / Qwen 72B & 109 & [39.4, 299] & \textbf{< .001} \\
 & LLaVA 34B / Qwen 235B & 1260 & [77.8, 20400] & \textbf{< .001} \\
 & LLaVA 34B / InternVL 38B & 152 & [48.9, 471] & \textbf{< .001} \\
 & Qwen 72B / Qwen 235B & 11.6 & [0.668, 202] & .132 \\
 & Qwen 72B / InternVL 38B & 1.40 & [0.381, 5.12] & .956 \\
 & Qwen 235B / InternVL 38B & 0.120 & [0.007, 2.19] & .270 \\
\midrule
40\% & Human / LLaVA 34B & 0.105 & [0.020, 0.559] & \textbf{.002} \\
 & Human / Qwen 72B & 5.49 & [0.970, 31.0] & .057 \\
 & Human / Qwen 235B & 33.6 & [3.56, 317] & \textbf{< .001} \\
 & Human / InternVL 38B & 9.70 & [1.57, 59.8] & \textbf{.006} \\
 & LLaVA 34B / Qwen 72B & 52.1 & [22.0, 123] & \textbf{< .001} \\
 & LLaVA 34B / Qwen 235B & 318 & [60.2, 1680] & \textbf{< .001} \\
 & LLaVA 34B / InternVL 38B & 92.0 & [33.1, 256] & \textbf{< .001} \\
 & Qwen 72B / Qwen 235B & 6.11 & [1.08, 34.5] & \textbf{.035} \\
 & Qwen 72B / InternVL 38B & 1.77 & [0.574, 5.44] & .640 \\
 & Qwen 235B / InternVL 38B & 0.289 & [0.047, 1.78] & .337 \\
\midrule
70\% & Human / LLaVA 34B & 0.072 & [0.015, 0.349] & \textbf{< .001} \\
 & Human / Qwen 72B & 1.34 & [0.276, 6.51] & .987 \\
 & Human / Qwen 235B & 20.1 & [2.27, 178] & \textbf{.002} \\
 & Human / InternVL 38B & 1.18 & [0.244, 5.70] & .999 \\
 & LLaVA 34B / Qwen 72B & 18.6 & [9.24, 37.4] & \textbf{< .001} \\
 & LLaVA 34B / Qwen 235B & 279 & [53.0, 1460] & \textbf{< .001} \\
 & LLaVA 34B / InternVL 38B & 16.3 & [8.24, 32.4] & \textbf{< .001} \\
 & Qwen 72B / Qwen 235B & 15.0 & [2.84, 79.0] & \textbf{< .001} \\
 & Qwen 72B / InternVL 38B & 0.880 & [0.441, 1.76] & .987 \\
 & Qwen 235B / InternVL 38B & 0.059 & [0.011, 0.308] & \textbf{< .001} \\
\bottomrule
\end{tabular}
}
\caption{Tukey-adjusted pairwise comparisons for \textit{associativeness} by noise level in PixCC. Estimates are odds ratios of an associative description (first relative to second source; $>1$: first source more associative). CIs are Tukey-adjusted and back-transformed from the log-odds scale. Odds ratios are conditional on a typical participant.}
\label{tab:pairs_assoc_pixcc}
\end{table}

\begin{figure}
    \centering
    \includegraphics[width=1\linewidth]{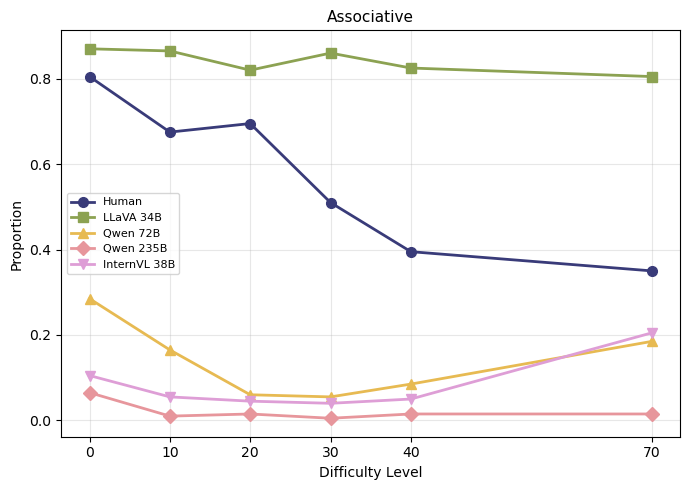}
    \caption{Associativity on all Data (PixCC)}
    \label{fig:succ_asso_full}
\end{figure}

\subsection{Effectiveness}


\begin{table}[H]
\scriptsize
\setlength{\tabcolsep}{1.5pt}
\centering
\resizebox{0.5\textwidth}{!}{%
\begin{tabular}{lrrrrl}
\toprule
\textbf{Coefficients} & \textbf{Estimate} & \textbf{Std. Error} & \textbf{z} & \textbf{OR (95\% CI)} & \textbf{p} \\
\midrule
InternVL 38B                      &  1.37756 & 0.06672 &  20.646 & 3.97 [3.48, 4.52] & \textbf{< .001} \\
LLaVA 34B                         & -1.04433 & 0.03506 & -29.783 & 0.35 [0.33, 0.38] & \textbf{< .001} \\
Qwen 235B                         & -1.50122 & 0.03887 & -38.619 & 0.22 [0.21, 0.24] & \textbf{< .001} \\
Qwen 72B                          & -0.26512 & 0.03640 &  -7.283 & 0.77 [0.71, 0.82] & \textbf{< .001} \\
difficulty\_level.L               & -3.99126 & 0.39184 & -10.186 & 0.02 [0.01, 0.04] & \textbf{< .001} \\
difficulty\_level.Q               &  1.20349 & 0.38539 &   3.123 & 3.33 [1.57, 7.09] & \textbf{.002} \\
difficulty\_level.C               &  0.56143 & 0.37903 &   1.481 & 1.75 [0.83, 3.69] & .139 \\
InternVL 38B:difficulty\_level.L  & -0.57286 & 0.12989 &  -4.410 & 0.56 [0.44, 0.73] & \textbf{< .001} \\
LLaVA 34B:difficulty\_level.L     &  0.61458 & 0.07424 &   8.278 & 1.85 [1.60, 2.14] & \textbf{< .001} \\
Qwen 235B:difficulty\_level.L     &  1.29414 & 0.08123 &  15.932 & 3.65 [3.11, 4.28] & \textbf{< .001} \\
Qwen 72B:difficulty\_level.L      &  0.38658 & 0.07699 &   5.021 & 1.47 [1.27, 1.71] & \textbf{< .001} \\
InternVL 38B:difficulty\_level.Q  &  0.43258 & 0.09581 &   4.515 & 1.54 [1.28, 1.86] & \textbf{< .001} \\
LLaVA 34B:difficulty\_level.Q     & -0.03372 & 0.06573 &  -0.513 & 0.97 [0.85, 1.10] & .608 \\
Qwen 235B:difficulty\_level.Q     &  0.30987 & 0.07340 &   4.222 & 1.36 [1.18, 1.57] & \textbf{< .001} \\
Qwen 72B:difficulty\_level.Q      & -0.46580 & 0.06910 &  -6.741 & 0.63 [0.55, 0.72] & \textbf{< .001} \\
InternVL 38B:difficulty\_level.C  & -0.05195 & 0.07286 &  -0.713 & 0.95 [0.82, 1.10] & .476 \\
LLaVA 34B:difficulty\_level.C     & -0.23855 & 0.05826 &  -4.095 & 0.79 [0.70, 0.88] & \textbf{< .001} \\
Qwen 235B:difficulty\_level.C     & -1.16962 & 0.06699 & -17.459 & 0.31 [0.27, 0.35] & \textbf{< .001} \\
Qwen 72B:difficulty\_level.C      &  0.49698 & 0.06208 &   8.005 & 1.64 [1.46, 1.86] & \textbf{< .001} \\
\bottomrule
\end{tabular}
}
\caption{Cumulative Link Mixed Model for \textit{effectiveness} in SilCC
$\variable{reflects} \sim \variable{source} * \variable{difficulty\_level}
+ (0 + \variable{is\_human} + \variable{human\_diff} \mid \variable{source\_id})
+ (1 \mid \variable{shapeID})$}
\label{tab:clmm_successfulness}
\end{table}

\begin{table}[H]
\scriptsize
\setlength{\tabcolsep}{4pt}
\centering
\begin{tabular}{lrrr}
\toprule
    \textbf{Effect} & $\boldsymbol{\chi^2}$ & \textbf{df} & \textbf{p} \\
\midrule
    source                         & 2149.812 &  4 & \textbf{< .001} \\
    difficulty\_level              &  111.099 &  3 & \textbf{< .001} \\
    source:difficulty\_level       &  844.932 & 12 & \textbf{< .001} \\
\bottomrule
\end{tabular}
\caption{Type III Wald $\chi^2$ tests for \textit{effectiveness} in SilCC. Cumulative link mixed model on the graded reflects label (no $<$ okay $<$ yes): $\variable{reflects} \sim \variable{source} * \variable{difficulty\_level}
+ (0 + \variable{is\_human} + \variable{human\_diff} \mid \variable{source\_id})
+ (1 \mid \variable{shapeID})$}
\label{tab:anova_mr_g_clmm}
\end{table}

\begin{table}[H]
\scriptsize
\setlength{\tabcolsep}{4pt}
\centering
\resizebox{0.5\textwidth}{!}{%
\begin{tabular}{llrrl}
\toprule
\textbf{Difficulty} & \textbf{Contrast} & \textbf{Estimate} & \textbf{95\% CI} & \textbf{p} \\
\midrule
A & Human / InternVL 38B &  3.410 & [ 2.777,  4.043] & \textbf{< .001} \\
  & Human / LLaVA 34B    &  3.943 & [ 3.290,  4.596] & \textbf{< .001} \\
  & Human / Qwen 72B     &  2.858 & [ 2.218,  3.498] & \textbf{< .001} \\
  & Human / Qwen 235B    & -0.262 & [-1.085,  0.561] & .909 \\
  & InternVL 38B / LLaVA 34B &  0.533 & [ 0.178,  0.888] & \textbf{< .001} \\
  & InternVL 38B / Qwen 72B  & -0.552 & [-0.907, -0.197] & \textbf{< .001} \\
  & InternVL 38B / Qwen 235B & -3.672 & [-4.316, -3.027] & \textbf{< .001} \\
  & LLaVA 34B / Qwen 72B     & -1.084 & [-1.467, -0.702] & \textbf{< .001} \\
  & LLaVA 34B / Qwen 235B    & -4.205 & [-4.869, -3.541] & \textbf{< .001} \\
  & Qwen 72B / Qwen 235B     & -3.120 & [-3.770, -2.470] & \textbf{< .001} \\
\midrule
B & Human / InternVL 38B &  2.579 & [ 2.215,  2.944] & \textbf{< .001} \\
  & Human / LLaVA 34B    &  3.985 & [ 3.588,  4.382] & \textbf{< .001} \\
  & Human / Qwen 72B     &  1.040 & [ 0.665,  1.415] & \textbf{< .001} \\
  & Human / Qwen 235B    & -1.331 & [-1.828, -0.835] & \textbf{< .001} \\
  & InternVL 38B / LLaVA 34B &  1.405 & [ 1.124,  1.687] & \textbf{< .001} \\
  & InternVL 38B / Qwen 72B  & -1.540 & [-1.827, -1.252] & \textbf{< .001} \\
  & InternVL 38B / Qwen 235B & -3.911 & [-4.355, -3.467] & \textbf{< .001} \\
  & LLaVA 34B / Qwen 72B     & -2.945 & [-3.269, -2.621] & \textbf{< .001} \\
  & LLaVA 34B / Qwen 235B    & -5.316 & [-5.789, -4.843] & \textbf{< .001} \\
  & Qwen 72B / Qwen 235B     & -2.371 & [-2.821, -1.921] & \textbf{< .001} \\
\midrule
C & Human / InternVL 38B &  1.798 & [ 1.570,  2.026] & \textbf{< .001} \\
  & Human / LLaVA 34B    &  1.650 & [ 1.415,  1.885] & \textbf{< .001} \\
  & Human / Qwen 72B     &  1.347 & [ 1.124,  1.571] & \textbf{< .001} \\
  & Human / Qwen 235B    &  0.545 & [ 0.320,  0.770] & \textbf{< .001} \\
  & InternVL 38B / LLaVA 34B & -0.148 & [-0.335,  0.040] & .199 \\
  & InternVL 38B / Qwen 72B  & -0.451 & [-0.624, -0.278] & \textbf{< .001} \\
  & InternVL 38B / Qwen 235B & -1.253 & [-1.432, -1.075] & \textbf{< .001} \\
  & LLaVA 34B / Qwen 72B     & -0.303 & [-0.486, -0.120] & \textbf{< .001} \\
  & LLaVA 34B / Qwen 235B    & -1.106 & [-1.293, -0.918] & \textbf{< .001} \\
  & Qwen 72B / Qwen 235B     & -0.803 & [-0.975, -0.630] & \textbf{< .001} \\
\midrule
D & Human / InternVL 38B &  1.900 & [ 1.658,  2.142] & \textbf{< .001} \\
  & Human / LLaVA 34B    &  1.938 & [ 1.690,  2.186] & \textbf{< .001} \\
  & Human / Qwen 72B     &  1.326 & [ 1.088,  1.563] & \textbf{< .001} \\
  & Human / Qwen 235B    &  0.826 & [ 0.591,  1.062] & \textbf{< .001} \\
  & InternVL 38B / LLaVA 34B &  0.037 & [-0.161,  0.236] & .986 \\
  & InternVL 38B / Qwen 72B  & -0.575 & [-0.763, -0.387] & \textbf{< .001} \\
  & InternVL 38B / Qwen 235B & -1.074 & [-1.260, -0.888] & \textbf{< .001} \\
  & LLaVA 34B / Qwen 72B     & -0.612 & [-0.808, -0.417] & \textbf{< .001} \\
  & LLaVA 34B / Qwen 235B    & -1.111 & [-1.305, -0.918] & \textbf{< .001} \\
  & Qwen 72B / Qwen 235B     & -0.499 & [-0.680, -0.318] & \textbf{< .001} \\
\bottomrule
\end{tabular}
}
\caption{Tukey-adjusted pairwise comparisons for effectiveness by difficulty level in SilCC. Estimates are differences between the first and second source; positive values indicate a higher estimated effectiveness for the first source.}
\label{tab:pairs_effectiveness}
\end{table}

\begin{table}[H]
\scriptsize
\setlength{\tabcolsep}{1.5pt}
\centering
\resizebox{0.5\textwidth}{!}{%
\begin{tabular}{lrrrll}
\toprule
\textbf{Coefficients} & \textbf{Estimate} & \textbf{Std. Error} & \textbf{z} & \textbf{OR (95\% CI)} & \textbf{p} \\
\midrule
Human & 1.76944 & 0.15175 & 11.660 & 5.87 [4.36, 7.90] & \textbf{< .001} \\
LLaVA 34B & -0.19099 & 0.07635 & -2.502 & 0.83 [0.71, 0.96] & \textbf{.012} \\
Qwen 72B & -0.39592 & 0.07658 & -5.170 & 0.67 [0.58, 0.78] & \textbf{< .001} \\
Qwen 235B & 1.96963 & 0.10986 & 17.928 & 7.17 [5.78, 8.89] & \textbf{< .001} \\
noise\_level.L & -0.28540 & 0.18299 & -1.560 & 0.75 [0.53, 1.08] & .119 \\
noise\_level.Q & -0.13177 & 0.14217 & -0.927 & 0.88 [0.66, 1.16] & .354 \\
noise\_level.C & 0.10900 & 0.10412 & 1.047 & 1.12 [0.91, 1.37] & .295 \\
noise\_level\textasciicircum 4 & -0.02711 & 0.09226 & -0.294 & 0.97 [0.81, 1.17] & .769 \\
noise\_level\textasciicircum 5 & -0.24542 & 0.09186 & -2.672 & 0.78 [0.65, 0.94] & \textbf{.008} \\
Human:noise\_level.L & -2.53389 & 0.28869 & -8.777 & 0.08 [0.05, 0.14] & \textbf{< .001} \\
LLaVA 34B:noise\_level.L & 2.25540 & 0.22726 & 9.924 & 9.54 [6.11, 14.89] & \textbf{< .001} \\
Qwen 72B:noise\_level.L & 2.01711 & 0.23588 & 8.551 & 7.52 [4.73, 11.93] & \textbf{< .001} \\
Qwen 235B:noise\_level.L & 1.47181 & 0.31385 & 4.690 & 4.36 [2.36, 8.06] & \textbf{< .001} \\
Human:noise\_level.Q & 0.65396 & 0.27150 & 2.409 & 1.92 [1.13, 3.27] & \textbf{.016} \\
LLaVA 34B:noise\_level.Q & 0.05865 & 0.18880 & 0.311 & 1.06 [0.73, 1.54] & .756 \\
Qwen 72B:noise\_level.Q & 1.08985 & 0.19025 & 5.729 & 2.97 [2.05, 4.32] & \textbf{< .001} \\
Qwen 235B:noise\_level.Q & -0.07578 & 0.27385 & -0.277 & 0.93 [0.54, 1.59] & .782 \\
Human:noise\_level.C & 0.06036 & 0.25882 & 0.233 & 1.06 [0.64, 1.76] & .816 \\
LLaVA 34B:noise\_level.C & -1.00339 & 0.15849 & -6.331 & 0.37 [0.27, 0.50] & \textbf{< .001} \\
Qwen 72B:noise\_level.C & 1.13751 & 0.14961 & 7.603 & 3.12 [2.33, 4.18] & \textbf{< .001} \\
Qwen 235B:noise\_level.C & -0.22403 & 0.24370 & -0.919 & 0.80 [0.50, 1.29] & .358 \\
Human:noise\_level\textasciicircum 4 & 0.12312 & 0.22047 & 0.558 & 1.13 [0.73, 1.74] & .577 \\
LLaVA 34B:noise\_level\textasciicircum 4 & -0.28892 & 0.14480 & -1.995 & 0.75 [0.56, 0.99] & \textbf{.046} \\
Qwen 72B:noise\_level\textasciicircum 4 & 0.48962 & 0.13911 & 3.520 & 1.63 [1.24, 2.14] & \textbf{< .001} \\
Qwen 235B:noise\_level\textasciicircum 4 & -0.16384 & 0.23173 & -0.707 & 0.85 [0.54, 1.34] & .480 \\
Human:noise\_level\textasciicircum 5 & -0.35875 & 0.23285 & -1.541 & 0.70 [0.44, 1.10] & .123 \\
LLaVA 34B:noise\_level\textasciicircum 5 & 0.35545 & 0.13850 & 2.566 & 1.43 [1.09, 1.87] & \textbf{.010} \\
Qwen 72B:noise\_level\textasciicircum 5 & -0.17262 & 0.14123 & -1.222 & 0.84 [0.64, 1.11] & .222 \\
Qwen 235B:noise\_level\textasciicircum 5 & 0.23089 & 0.22435 & 1.029 & 1.26 [0.81, 1.96] & .303 \\
\bottomrule
\end{tabular}
}
\caption{Cumulative link mixed model for \textit{effectiveness} in PixCC: $\variable{reflects} \sim \variable{source} * \variable{noise\_level} + (0 + \variable{is\_human} \mid \variable{participant}) + (1 \mid \variable{image})$.}
\label{tab:coef_eff_pixcc}
\end{table}

\begin{table}[H]
\scriptsize
\setlength{\tabcolsep}{6pt}
\centering
\begin{tabular}{lrrl}
\toprule
\textbf{Effect} & $\boldsymbol{\chi^2}$ & \textbf{df} & \textbf{p} \\
\midrule
source & 708.91 & 4 & \textbf{< .001} \\
noise\_level & 12.21 & 5 & \textbf{.032} \\
source:noise\_level & 482.04 & 20 & \textbf{< .001} \\
\bottomrule
\end{tabular}
\caption{Type III Wald $\chi^2$ tests for \textit{effectiveness} in PixCC. Cumulative link mixed model: $\variable{reflects} \sim \variable{source} * \variable{noise\_level} + (0 + \variable{is\_human} \mid \variable{participant}) + (1 \mid \variable{image})$}
\label{tab:anova_eff_pixcc}
\end{table}
 
\begin{table}[H]
\scriptsize
\setlength{\tabcolsep}{1.5pt}
\centering
\resizebox{0.5\textwidth}{!}{%
\begin{tabular}{llrrl}
\toprule
\textbf{Noise} & \textbf{Contrast} & \textbf{Estimate} & \textbf{95\% CI} & \textbf{p} \\
\midrule
0\% & Human -- LLaVA 34B & 4.418 & [3.045, 5.792] & \textbf{< .001} \\
 & Human -- Qwen 72B & 4.653 & [3.276, 6.030] & \textbf{< .001} \\
 & Human -- Qwen 235B & 2.240 & [0.797, 3.684] & \textbf{< .001} \\
 & Human -- InternVL 38B & 5.980 & [4.574, 7.387] & \textbf{< .001} \\
 & LLaVA 34B -- Qwen 72B & 0.235 & [-0.288, 0.757] & .736 \\
 & LLaVA 34B -- Qwen 235B & -2.178 & [-2.860, -1.496] & \textbf{< .001} \\
 & LLaVA 34B -- InternVL 38B & 1.562 & [0.968, 2.156] & \textbf{< .001} \\
 & Qwen 72B -- Qwen 235B & -2.413 & [-3.103, -1.723] & \textbf{< .001} \\
 & Qwen 72B -- InternVL 38B & 1.327 & [0.725, 1.930] & \textbf{< .001} \\
 & Qwen 235B -- InternVL 38B & 3.740 & [2.993, 4.487] & \textbf{< .001} \\
\midrule
10\% & Human -- LLaVA 34B & 3.515 & [2.531, 4.499] & \textbf{< .001} \\
 & Human -- Qwen 72B & 3.326 & [2.343, 4.309] & \textbf{< .001} \\
 & Human -- Qwen 235B & 0.892 & [-0.235, 2.019] & .196 \\
 & Human -- InternVL 38B & 4.584 & [3.571, 5.597] & \textbf{< .001} \\
 & LLaVA 34B -- Qwen 72B & -0.188 & [-0.699, 0.322] & .852 \\
 & LLaVA 34B -- Qwen 235B & -2.623 & [-3.377, -1.868] & \textbf{< .001} \\
 & LLaVA 34B -- InternVL 38B & 1.069 & [0.506, 1.632] & \textbf{< .001} \\
 & Qwen 72B -- Qwen 235B & -2.434 & [-3.187, -1.682] & \textbf{< .001} \\
 & Qwen 72B -- InternVL 38B & 1.258 & [0.697, 1.818] & \textbf{< .001} \\
 & Qwen 235B -- InternVL 38B & 3.692 & [2.901, 4.483] & \textbf{< .001} \\
\midrule
20\% & Human -- LLaVA 34B & 3.606 & [2.605, 4.607] & \textbf{< .001} \\
 & Human -- Qwen 72B & 2.506 & [1.493, 3.519] & \textbf{< .001} \\
 & Human -- Qwen 235B & 0.815 & [-0.350, 1.980] & .312 \\
 & Human -- InternVL 38B & 4.490 & [3.462, 5.518] & \textbf{< .001} \\
 & LLaVA 34B -- Qwen 72B & -1.100 & [-1.625, -0.575] & \textbf{< .001} \\
 & LLaVA 34B -- Qwen 235B & -2.791 & [-3.570, -2.011] & \textbf{< .001} \\
 & LLaVA 34B -- InternVL 38B & 0.884 & [0.336, 1.432] & \textbf{< .001} \\
 & Qwen 72B -- Qwen 235B & -1.691 & [-2.487, -0.895] & \textbf{< .001} \\
 & Qwen 72B -- InternVL 38B & 1.984 & [1.410, 2.558] & \textbf{< .001} \\
 & Qwen 235B -- InternVL 38B & 3.675 & [2.861, 4.489] & \textbf{< .001} \\
\midrule
30\% & Human -- LLaVA 34B & 1.387 & [0.648, 2.126] & \textbf{< .001} \\
 & Human -- Qwen 72B & 1.898 & [1.156, 2.639] & \textbf{< .001} \\
 & Human -- Qwen 235B & -0.809 & [-1.802, 0.184] & .171 \\
 & Human -- InternVL 38B & 3.910 & [3.092, 4.728] & \textbf{< .001} \\
 & LLaVA 34B -- Qwen 72B & 0.510 & [-0.012, 1.033] & .059 \\
 & LLaVA 34B -- Qwen 235B & -2.196 & [-3.039, -1.353] & \textbf{< .001} \\
 & LLaVA 34B -- InternVL 38B & 2.523 & [1.899, 3.147] & \textbf{< .001} \\
 & Qwen 72B -- Qwen 235B & -2.706 & [-3.551, -1.861] & \textbf{< .001} \\
 & Qwen 72B -- InternVL 38B & 2.012 & [1.387, 2.638] & \textbf{< .001} \\
 & Qwen 235B -- InternVL 38B & 4.719 & [3.806, 5.631] & \textbf{< .001} \\
\midrule
40\% & Human -- LLaVA 34B & 0.003 & [-0.778, 0.785] & 1.000 \\
 & Human -- Qwen 72B & 2.252 & [1.548, 2.956] & \textbf{< .001} \\
 & Human -- Qwen 235B & -1.545 & [-2.656, -0.435] & \textbf{.001} \\
 & Human -- InternVL 38B & 3.942 & [3.126, 4.758] & \textbf{< .001} \\
 & LLaVA 34B -- Qwen 72B & 2.248 & [1.632, 2.864] & \textbf{< .001} \\
 & LLaVA 34B -- Qwen 235B & -1.549 & [-2.605, -0.492] & \textbf{< .001} \\
 & LLaVA 34B -- InternVL 38B & 3.939 & [3.197, 4.680] & \textbf{< .001} \\
 & Qwen 72B -- Qwen 235B & -3.797 & [-4.798, -2.796] & \textbf{< .001} \\
 & Qwen 72B -- InternVL 38B & 1.691 & [1.037, 2.345] & \textbf{< .001} \\
 & Qwen 235B -- InternVL 38B & 5.488 & [4.404, 6.571] & \textbf{< .001} \\
\midrule
70\% & Human -- LLaVA 34B & -1.167 & [-1.902, -0.432] & \textbf{< .001} \\
 & Human -- Qwen 72B & -1.642 & [-2.439, -0.846] & \textbf{< .001} \\
 & Human -- Qwen 235B & -2.794 & [-3.961, -1.628] & \textbf{< .001} \\
 & Human -- InternVL 38B & 6.623 & [3.840, 9.406] & \textbf{< .001} \\
 & LLaVA 34B -- Qwen 72B & -0.476 & [-1.291, 0.340] & .503 \\
 & LLaVA 34B -- Qwen 235B & -1.628 & [-2.807, -0.448] & \textbf{.002} \\
 & LLaVA 34B -- InternVL 38B & 7.790 & [5.000, 10.579] & \textbf{< .001} \\
 & Qwen 72B -- Qwen 235B & -1.152 & [-2.370, 0.067] & .074 \\
 & Qwen 72B -- InternVL 38B & 8.265 & [5.459, 11.072] & \textbf{< .001} \\
 & Qwen 235B -- InternVL 38B & 9.417 & [6.485, 12.350] & \textbf{< .001} \\
\bottomrule
\end{tabular}
}
\caption{Tukey-adjusted pairwise comparisons for \textit{effectiveness} by noise level in PixCC. Estimates are differences in log-odds of a higher reflects rating (first minus second source); CIs are Tukey-adjusted.}
\label{tab:pairs_eff_pixcc}
\end{table}

\subsection{Novelty}
\label{app:nov}

\begin{table}[H]
\scriptsize
\setlength{\tabcolsep}{1.5pt}
\centering
\resizebox{0.5\textwidth}{!}{%
\begin{tabular}{lrrrrl}
\toprule
\textbf{Coefficients} & \textbf{Estimate} & \textbf{Std. Error} & \textbf{t} & \textbf{95\% CI} & \textbf{p} \\
\midrule
(Intercept)                      &  0.064439 & 0.002639 &  24.415 & [ 0.06,  0.07] & \textbf{< .001} \\
InternVL 38B                     &  0.094523 & 0.004528 &  20.874 & [ 0.09,  0.10] & \textbf{< .001} \\
LLaVA 34B                        & -0.050661 & 0.004528 & -11.188 & [-0.06, -0.04] & \textbf{< .001} \\
Qwen 235B                        &  0.026590 & 0.004544 &   5.852 & [ 0.02,  0.04] & \textbf{< .001} \\
Qwen 72B                         & -0.041895 & 0.004528 &  -9.252 & [-0.05, -0.03] & \textbf{< .001} \\
difficulty\_level.L              &  0.082332 & 0.005276 &  15.606 & [ 0.07,  0.09] & \textbf{< .001} \\
difficulty\_level.Q              & -0.001249 & 0.005279 &  -0.237 & [-0.01,  0.01] & .814 \\
difficulty\_level.C              & -0.016873 & 0.005281 &  -3.195 & [-0.03, -0.01] & \textbf{.002} \\
InternVL 38B:difficulty\_level.L &  0.099562 & 0.009055 &  10.995 & [ 0.08,  0.12] & \textbf{< .001} \\
LLaVA 34B:difficulty\_level.L    & -0.063497 & 0.009055 &  -7.012 & [-0.08, -0.05] & \textbf{< .001} \\
Qwen 235B:difficulty\_level.L    &  0.040603 & 0.009061 &   4.481 & [ 0.02,  0.06] & \textbf{< .001} \\
Qwen 72B:difficulty\_level.L     & -0.051131 & 0.009055 &  -5.647 & [-0.07, -0.03] & \textbf{< .001} \\
InternVL 38B:difficulty\_level.Q & -0.052092 & 0.009057 &  -5.752 & [-0.07, -0.03] & \textbf{< .001} \\
LLaVA 34B:difficulty\_level.Q    &  0.011186 & 0.009057 &   1.235 & [-0.01,  0.03] & .218 \\
Qwen 235B:difficulty\_level.Q    &  0.026450 & 0.009088 &   2.911 & [ 0.01,  0.04] & \textbf{.004} \\
Qwen 72B:difficulty\_level.Q     &  0.000018 & 0.009057 &   0.002 & [-0.02,  0.02] & .998 \\
InternVL 38B:difficulty\_level.C & -0.048984 & 0.009058 &  -5.408 & [-0.07, -0.03] & \textbf{< .001} \\
LLaVA 34B:difficulty\_level.C    &  0.015902 & 0.009058 &   1.756 & [-0.00,  0.03] & .080 \\
Qwen 235B:difficulty\_level.C    &  0.010283 & 0.009114 &   1.128 & [-0.01,  0.03] & .260 \\
Qwen 72B:difficulty\_level.C     &  0.021335 & 0.009058 &   2.355 & [ 0.00,  0.04] & \textbf{.019} \\
\bottomrule
\end{tabular}
}
\caption{Linear mixed model for \textit{novelty} in SilCC.
$\variable{novelty} \sim \variable{source} * \variable{difficulty\_level}
+ (1 \mid \variable{shapeID})$}
\label{tab:lmm_novelty}
\end{table}

\begin{table}[H]
\scriptsize
\setlength{\tabcolsep}{4pt}
\centering
\begin{tabular}{lrrr}
\toprule
    \textbf{Effect} & $\boldsymbol{\chi^2}$ & \textbf{df} & \textbf{p} \\
\midrule
    source                        & 576.61 &  4 & \textbf{< .001} \\
    difficulty\_level             & 254.04 &  3 & \textbf{< .001} \\
    source:difficulty\_level      & 255.37 & 12 & \textbf{< .001} \\
\bottomrule
\end{tabular}
\caption{Type III Wald $\chi^2$ tests for \textit{novelty} in SilCC. Linear mixed model for \textit{novelty}
$\variable{novelty} \sim \variable{source} * \variable{difficulty\_level}
+ (1 \mid \variable{shapeID})$}
\label{tab:anova_novelty}
\end{table}

\begin{table}[H]
\scriptsize
\setlength{\tabcolsep}{4pt}
\centering
\resizebox{0.5\textwidth}{!}{%
\begin{tabular}{llrrl}
\toprule
\textbf{Difficulty} & \textbf{Contrast} & \textbf{Estimate} & \textbf{95\% CI} & \textbf{p} \\
\midrule
A & Human / InternVL 38B &  0.019 & [-0.024,  0.062] & .757 \\
  & Human / LLaVA 34B    &  0.002 & [-0.041,  0.045] & 1.000 \\
  & Human / Qwen 72B     &  0.025 & [-0.018,  0.068] & .502 \\
  & Human / Qwen 235B    &  0.017 & [-0.026,  0.060] & .809 \\
  & InternVL 38B / LLaVA 34B & -0.016 & [-0.059,  0.027] & .837 \\
  & InternVL 38B / Qwen 72B  &  0.006 & [-0.037,  0.049] & .994 \\
  & InternVL 38B / Qwen 235B & -0.001 & [-0.045,  0.042] & 1.000 \\
  & LLaVA 34B / Qwen 72B     &  0.023 & [-0.020,  0.066] & .600 \\
  & LLaVA 34B / Qwen 235B    &  0.015 & [-0.028,  0.058] & .879 \\
  & Qwen 72B / Qwen 235B     & -0.008 & [-0.051,  0.035] & .987 \\
\midrule
B & Human / InternVL 38B &  0.097 & [ 0.054,  0.140] & \textbf{< .001} \\
  & Human / LLaVA 34B    &  0.054 & [ 0.011,  0.097] & \textbf{.006} \\
  & Human / Qwen 72B     &  0.082 & [ 0.039,  0.125] & \textbf{< .001} \\
  & Human / Qwen 235B    &  0.095 & [ 0.051,  0.138] & \textbf{< .001} \\
  & InternVL 38B / LLaVA 34B & -0.043 & [-0.086,  0.000] & .054 \\
  & InternVL 38B / Qwen 72B  & -0.015 & [-0.058,  0.028] & .868 \\
  & InternVL 38B / Qwen 235B & -0.002 & [-0.045,  0.041] & 1.000 \\
  & LLaVA 34B / Qwen 72B     &  0.027 & [-0.016,  0.070] & .409 \\
  & LLaVA 34B / Qwen 235B    &  0.040 & [-0.003,  0.083] & .079 \\
  & Qwen 72B / Qwen 235B     &  0.013 & [-0.030,  0.056] & .923 \\
\midrule
C & Human / InternVL 38B &  0.257 & [ 0.222,  0.292] & \textbf{< .001} \\
  & Human / LLaVA 34B    &  0.160 & [ 0.125,  0.196] & \textbf{< .001} \\
  & Human / Qwen 72B     &  0.243 & [ 0.208,  0.278] & \textbf{< .001} \\
  & Human / Qwen 235B    &  0.218 & [ 0.183,  0.253] & \textbf{< .001} \\
  & InternVL 38B / LLaVA 34B & -0.097 & [-0.132, -0.061] & \textbf{< .001} \\
  & InternVL 38B / Qwen 72B  & -0.013 & [-0.049,  0.022] & .831 \\
  & InternVL 38B / Qwen 235B & -0.039 & [-0.074, -0.004] & \textbf{.023} \\
  & LLaVA 34B / Qwen 72B     &  0.083 & [ 0.048,  0.119] & \textbf{< .001} \\
  & LLaVA 34B / Qwen 235B    &  0.058 & [ 0.022,  0.094] & \textbf{.0001} \\
  & Qwen 72B / Qwen 235B     & -0.025 & [-0.060,  0.010] & .285 \\
\midrule
D & Human / InternVL 38B &  0.208 & [ 0.173,  0.244] & \textbf{< .001} \\
  & Human / LLaVA 34B    &  0.055 & [ 0.020,  0.090] & \textbf{< .001} \\
  & Human / Qwen 72B     &  0.196 & [ 0.161,  0.231] & \textbf{< .001} \\
  & Human / Qwen 235B    &  0.162 & [ 0.127,  0.198] & \textbf{< .001} \\
  & InternVL 38B / LLaVA 34B & -0.153 & [-0.189, -0.118] & \textbf{< .001} \\
  & InternVL 38B / Qwen 72B  & -0.013 & [-0.048,  0.022] & .859 \\
  & InternVL 38B / Qwen 235B & -0.046 & [-0.081, -0.011] & \textbf{.004} \\
  & LLaVA 34B / Qwen 72B     &  0.141 & [ 0.106,  0.176] & \textbf{< .001} \\
  & LLaVA 34B / Qwen 235B    &  0.107 & [ 0.072,  0.143] & \textbf{< .001} \\
  & Qwen 72B / Qwen 235B     & -0.033 & [-0.068,  0.002] & .073 \\
\bottomrule
\end{tabular}
}
\caption{Tukey-adjusted pairwise comparisons for novelty by difficulty level in SilCC. Estimates are differences between the first and second source; positive values indicate a higher estimated mean for the first source.}
\label{tab:pairs_by_difficulty_novelty}
\end{table}

\begin{table}[H]
\scriptsize
\setlength{\tabcolsep}{1.5pt}
\centering
\resizebox{0.5\textwidth}{!}{%
\begin{tabular}{lrrrll}
\toprule
\textbf{Coefficients} & \textbf{Estimate} & \textbf{Std. Error} & \textbf{t} & \textbf{95\% CI} & \textbf{p} \\
\midrule
(Intercept) & 0.36700 & 0.01355 & 27.075 & [0.34, 0.39] & \textbf{< .001} \\
Human & 0.08400 & 0.01283 & 6.546 & [0.06, 0.11] & \textbf{< .001} \\
noise\_level.L & 0.09687 & 0.02869 & 3.376 & [0.04, 0.15] & \textbf{< .001} \\
noise\_level.Q & -0.04675 & 0.02869 & -1.629 & [-0.10, 0.01] & .105 \\
noise\_level.C & 0.01187 & 0.02869 & 0.414 & [-0.04, 0.07] & .680 \\
noise\_level\textasciicircum 4 & 0.04946 & 0.02869 & 1.724 & [-0.01, 0.11] & .087 \\
Human:noise\_level.L & 0.09817 & 0.02869 & 3.421 & [0.04, 0.15] & \textbf{< .001} \\
Human:noise\_level.Q & -0.06763 & 0.02869 & -2.357 & [-0.12, -0.01] & \textbf{.019} \\
Human:noise\_level.C & -0.02551 & 0.02869 & -0.889 & [-0.08, 0.03] & .375 \\
Human:noise\_level\textasciicircum 4 & 0.02309 & 0.02869 & 0.805 & [-0.03, 0.08] & .422 \\
\bottomrule
\end{tabular}
}
\caption{Linear mixed model for \textit{weighted novelty} in PixCC (humans vs.\ LLaVA 34B, noise levels 10--70), fitted on one value per image $\times$ source cell: $\variable{novelty} \sim \variable{source} * \variable{noise\_level} + (1 \mid \variable{image})$.}
\label{tab:coef_novelty_pixcc}
\end{table}

\begin{table}[H]
\scriptsize
\setlength{\tabcolsep}{6pt}
\centering
\begin{tabular}{lrrl}
\toprule
\textbf{Effect} & $\boldsymbol{\chi^2}$ & \textbf{df} & \textbf{p} \\
\midrule
source & 42.86 & 1 & \textbf{< .001} \\
noise\_level & 17.20 & 4 & \textbf{.002} \\
source:noise\_level & 18.70 & 4 & \textbf{< .001} \\
\bottomrule
\end{tabular}
\caption{Type III Wald $\chi^2$ tests for weighted \textit{novelty} in PixCC (humans and LLaVA 34B, noise levels 10--70\%). Linear mixed model on one value per image and source: $\variable{novelty} \sim \variable{source} * \variable{noise\_level} + (1 \mid \variable{image})$.}
\label{tab:anova_novelty_pixcc}
\end{table}
 
\begin{table}[H]
\scriptsize
\setlength{\tabcolsep}{1.5pt}
\centering
\resizebox{0.5\textwidth}{!}{%
\begin{tabular}{llrrl}
\toprule
\textbf{Noise} & \textbf{Contrast} & \textbf{Estimate} & \textbf{95\% CI} & \textbf{p} \\
\midrule
10\% & Human -- LLaVA 34B & 0.015 & [-0.098, 0.128] & .794 \\
\midrule
20\% & Human -- LLaVA 34B & 0.055 & [-0.058, 0.168] & .339 \\
\midrule
30\% & Human -- LLaVA 34B & 0.225 & [0.112, 0.338] & \textbf{< .001} \\
\midrule
40\% & Human -- LLaVA 34B & 0.285 & [0.172, 0.398] & \textbf{< .001} \\
\midrule
70\% & Human -- LLaVA 34B & 0.260 & [0.147, 0.373] & \textbf{< .001} \\
\bottomrule
\end{tabular}
}
\caption{Pairwise comparisons for weighted \textit{novelty} by noise level in PixCC (Human minus LLaVA 34B; $SE = 0.057$, $df = 171$).}
\label{tab:pairs_novelty_pixcc}
\end{table}

\subsection{Semantic Similarity}

\begin{figure}[H]
    \centering
    \includegraphics[width=1\linewidth]{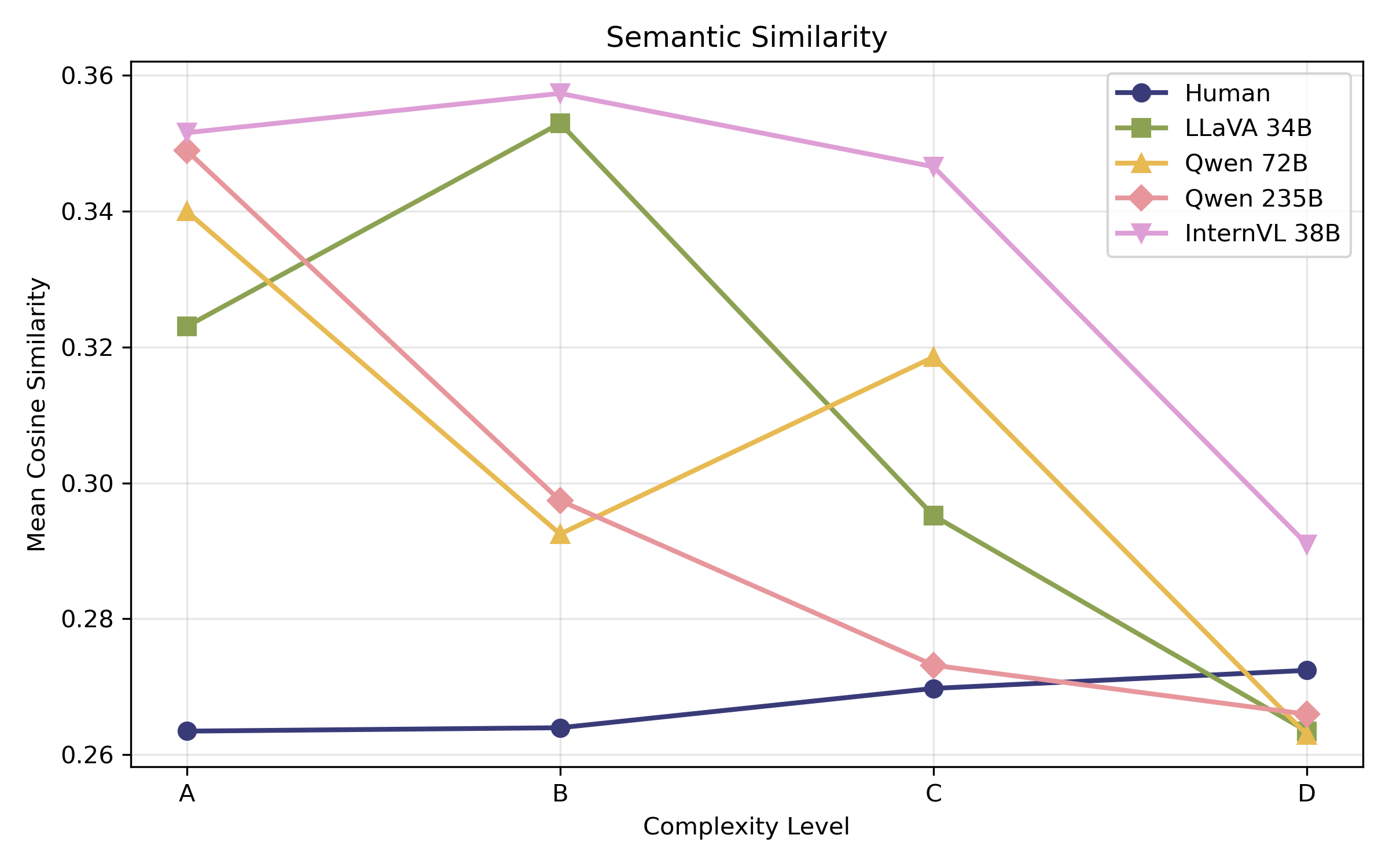}
    \caption{Semantic similarity for SilCC illustrating the average human and model responses across difficulty levels. }
    \label{fig:c02_semantic_sim}
\end{figure}

\begin{figure}[H]
    \centering
    \includegraphics[width=1\linewidth]{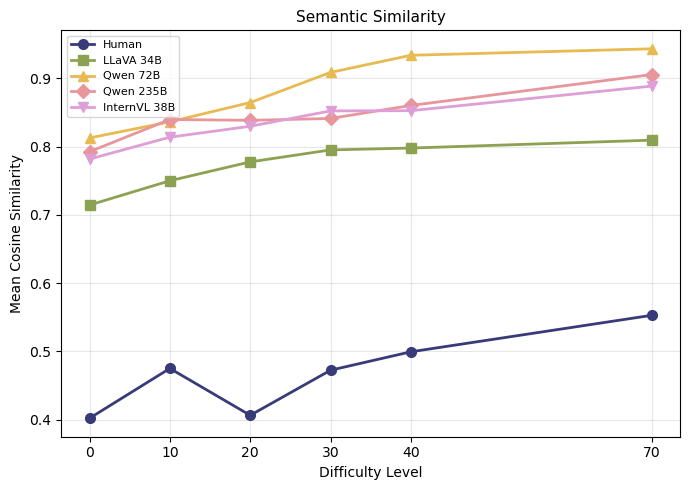}
    \caption{Semantic similarity for PixCC illustrating the average human and model responses across difficulty levels (on the full descriptions).}
    \label{fig:B02_semantic_sim}
\end{figure}

\subsection{Entropy}
\label{app:ent}

\begin{table}[H]
\scriptsize
\setlength{\tabcolsep}{1.5pt}
\centering
\resizebox{0.5\textwidth}{!}{%
\begin{tabular}{lrrrrl}
\toprule
\textbf{Coefficients} & \textbf{Estimate} & \textbf{Std. Error} & \textbf{t} & \textbf{95\% CI} & \textbf{p} \\
\midrule
(Intercept)                       &  2.19404 & 0.08352 &  26.268 & [ 2.03,  2.36] & \textbf{< .001} \\
InternVL 38B                      &  1.65111 & 0.09747 &  16.940 & [ 1.46,  1.84] & \textbf{< .001} \\
LLaVA 34B                         & -0.61332 & 0.09747 &  -6.292 & [-0.80, -0.42] & \textbf{< .001} \\
Qwen 235B                         &  0.38589 & 0.09784 &   3.944 & [ 0.19,  0.58] & \textbf{< .001} \\
Qwen 72B                          & -0.70489 & 0.09747 &  -7.232 & [-0.90, -0.51] & \textbf{< .001} \\
difficulty\_level.L               &  2.07201 & 0.16700 &  12.407 & [ 1.74,  2.40] & \textbf{< .001} \\
difficulty\_level.Q               & -0.01522 & 0.16705 &  -0.091 & [-0.34,  0.31] & .928 \\
difficulty\_level.C               & -0.09596 & 0.16709 &  -0.574 & [-0.42,  0.23] & .567 \\
InternVL 38B:difficulty\_level.L  &  0.97028 & 0.19490 &   4.978 & [ 0.59,  1.35] & \textbf{< .001} \\
LLaVA 34B:difficulty\_level.L     & -0.60494 & 0.19490 &  -3.104 & [-0.99, -0.22] & \textbf{.002} \\
Qwen 235B:difficulty\_level.L     &  0.06080 & 0.19505 &   0.312 & [-0.32,  0.44] & .755 \\
Qwen 72B:difficulty\_level.L      & -0.20165 & 0.19490 &  -1.035 & [-0.58,  0.18] & .302 \\
InternVL 38B:difficulty\_level.Q  & -0.78190 & 0.19494 &  -4.011 & [-1.17, -0.40] & \textbf{< .001} \\
LLaVA 34B:difficulty\_level.Q     &  0.18981 & 0.19494 &   0.974 & [-0.19,  0.57] & .331 \\
Qwen 235B:difficulty\_level.Q     &  0.47933 & 0.19567 &   2.450 & [ 0.09,  0.86] & \textbf{.015} \\
Qwen 72B:difficulty\_level.Q      &  0.03913 & 0.19494 &   0.201 & [-0.34,  0.42] & .841 \\
InternVL 38B:difficulty\_level.C  & -0.21906 & 0.19498 &  -1.123 & [-0.60,  0.16] & .262 \\
LLaVA 34B:difficulty\_level.C     & -0.07154 & 0.19498 &  -0.367 & [-0.45,  0.31] & .714 \\
Qwen 235B:difficulty\_level.C     &  0.29750 & 0.19629 &   1.516 & [-0.09,  0.68] & .131 \\
Qwen 72B:difficulty\_level.C      &  0.11703 & 0.19498 &   0.600 & [-0.27,  0.50] & .549 \\
\bottomrule
\end{tabular}
}
\caption{Linear mixed model for \textit{entropy} in SilCC.
$\variable{entropy} \sim \variable{source} * \variable{difficulty\_level}
+ (1 \mid \variable{shapeID})$}
\label{tab:lmm_entropy}
\end{table}

\begin{table}[H]
\scriptsize
\setlength{\tabcolsep}{4pt}
\centering
\begin{tabular}{lrrr}
\toprule
    \textbf{Effect} & $\boldsymbol{\chi^2}$ & \textbf{df} & \textbf{p} \\
\midrule
    source                        & 359.175 &  4 & \textbf{< .001} \\
    difficulty\_level             & 155.143 &  3 & \textbf{< .001} \\
    source:difficulty\_level      &  51.474 & 12 & \textbf{< .001} \\
\bottomrule
\end{tabular}
\caption{Type III Wald $\chi^2$ tests for \textit{entropy} in SilCC. Linear mixed model for \textit{entropy}
$\variable{entropy} \sim \variable{source} * \variable{difficulty\_level}
+ (1 \mid \variable{shapeID})$}
\label{tab:anova_entropy}
\end{table}

\begin{table}[H]
\scriptsize
\setlength{\tabcolsep}{4pt}
\centering
\resizebox{0.5\textwidth}{!}{%
\begin{tabular}{llrrl}
\toprule
\textbf{Difficulty} & \textbf{Contrast} & \textbf{Estimate} & \textbf{95\% CI} & \textbf{p} \\
\midrule
A & Human / InternVL 38B &  0.755 & [-0.172,  1.681] & .169 \\
  & Human / LLaVA 34B    &  0.140 & [-0.786,  1.066] & .994 \\
  & Human / Qwen 72B     &  1.235 & [ 0.308,  2.161] & \textbf{.003} \\
  & Human / Qwen 235B    &  1.162 & [ 0.236,  2.088] & \textbf{.006} \\
  & InternVL 38B / LLaVA 34B & -0.615 & [-1.541,  0.312] & .363 \\
  & InternVL 38B / Qwen 72B  &  0.480 & [-0.447,  1.406] & .615 \\
  & InternVL 38B / Qwen 235B &  0.407 & [-0.519,  1.333] & .748 \\
  & LLaVA 34B / Qwen 72B     &  1.095 & [ 0.168,  2.021] & \textbf{.011} \\
  & LLaVA 34B / Qwen 235B    &  1.022 & [ 0.096,  1.948] & \textbf{.022} \\
  & Qwen 72B / Qwen 235B     & -0.073 & [-0.999,  0.854] & 1.000 \\
\midrule
B & Human / InternVL 38B &  2.299 & [ 1.373,  3.226] & \textbf{< .001} \\
  & Human / LLaVA 34B    &  1.346 & [ 0.420,  2.272] & \textbf{< .001} \\
  & Human / Qwen 72B     &  2.279 & [ 1.353,  3.205] & \textbf{< .001} \\
  & Human / Qwen 235B    &  2.467 & [ 1.540,  3.393] & \textbf{< .001} \\
  & InternVL 38B / LLaVA 34B & -0.953 & [-1.880, -0.027] & \textbf{.040} \\
  & InternVL 38B / Qwen 72B  & -0.020 & [-0.947,  0.906] & 1.000 \\
  & InternVL 38B / Qwen 235B &  0.168 & [-0.759,  1.094] & .988 \\
  & LLaVA 34B / Qwen 72B     &  0.933 & [ 0.007,  1.859] & \textbf{.047} \\
  & LLaVA 34B / Qwen 235B    &  1.121 & [ 0.194,  2.047] & \textbf{.009} \\
  & Qwen 72B / Qwen 235B     &  0.188 & [-0.739,  1.114] & .981 \\
\midrule
C & Human / InternVL 38B &  3.202 & [ 2.445,  3.958] & \textbf{< .001} \\
  & Human / LLaVA 34B    &  2.446 & [ 1.680,  3.212] & \textbf{< .001} \\
  & Human / Qwen 72B     &  3.254 & [ 2.498,  4.010] & \textbf{< .001} \\
  & Human / Qwen 235B    &  3.129 & [ 2.372,  3.885] & \textbf{< .001} \\
  & InternVL 38B / LLaVA 34B & -0.756 & [-1.522,  0.010] & .055 \\
  & InternVL 38B / Qwen 72B  &  0.053 & [-0.704,  0.809] & 1.000 \\
  & InternVL 38B / Qwen 235B & -0.073 & [-0.829,  0.684] & .999 \\
  & LLaVA 34B / Qwen 72B     &  0.808 & [ 0.043,  1.574] & \textbf{.033} \\
  & LLaVA 34B / Qwen 235B    &  0.683 & [-0.083,  1.449] & .106 \\
  & Qwen 72B / Qwen 235B     & -0.125 & [-0.882,  0.631] & .991 \\
\midrule
D & Human / InternVL 38B &  2.802 & [ 2.046,  3.559] & \textbf{< .001} \\
  & Human / LLaVA 34B    &  1.129 & [ 0.373,  1.886] & \textbf{< .001} \\
  & Human / Qwen 72B     &  2.657 & [ 1.900,  3.413] & \textbf{< .001} \\
  & Human / Qwen 235B    &  2.722 & [ 1.966,  3.479] & \textbf{< .001} \\
  & InternVL 38B / LLaVA 34B & -1.673 & [-2.430, -0.917] & \textbf{< .001} \\
  & InternVL 38B / Qwen 72B  & -0.146 & [-0.902,  0.611] & .984 \\
  & InternVL 38B / Qwen 235B & -0.080 & [-0.836,  0.676] & .998 \\
  & LLaVA 34B / Qwen 72B     &  1.527 & [ 0.771,  2.284] & \textbf{< .001} \\
  & LLaVA 34B / Qwen 235B    &  1.593 & [ 0.837,  2.350] & \textbf{< .001} \\
  & Qwen 72B / Qwen 235B     &  0.066 & [-0.691,  0.822] & .999 \\
\bottomrule
\end{tabular}
}
\caption{Tukey-adjusted pairwise comparisons for entropy by difficulty level in SilCC. Estimates are differences between the first and second source; positive values indicate a higher estimated mean for the first source.}
\label{tab:pairs_by_difficulty}
\end{table}


\begin{table}[H]
\scriptsize
\setlength{\tabcolsep}{1.5pt}
\centering
\resizebox{0.5\textwidth}{!}{%
\begin{tabular}{lrrrll}
\toprule
\textbf{Coefficients} & \textbf{Estimate} & \textbf{Std. Error} & \textbf{t} & \textbf{95\% CI} & \textbf{p} \\
\midrule
(Intercept) & 0.93905 & 0.02010 & 46.728 & [0.90, 0.98] & \textbf{< .001} \\
Human & 1.58553 & 0.03202 & 49.523 & [1.52, 1.65] & \textbf{< .001} \\
LLaVA 34B & 1.15853 & 0.03202 & 36.186 & [1.10, 1.22] & \textbf{< .001} \\
Qwen 72B & -0.91573 & 0.03202 & -28.602 & [-0.98, -0.85] & \textbf{< .001} \\
Qwen 235B & -0.92355 & 0.03202 & -28.846 & [-0.99, -0.86] & \textbf{< .001} \\
noise\_level.L & 0.03986 & 0.03921 & 1.016 & [-0.04, 0.12] & .310 \\
noise\_level.Q & -0.03773 & 0.03921 & -0.962 & [-0.11, 0.04] & .336 \\
noise\_level.C & -0.02067 & 0.03921 & -0.527 & [-0.10, 0.06] & .598 \\
noise\_level\textasciicircum 4 & -0.00056 & 0.03921 & -0.014 & [-0.08, 0.08] & .989 \\
noise\_level\textasciicircum 5 & 0.02541 & 0.03921 & 0.648 & [-0.05, 0.10] & .517 \\
Human:noise\_level.L & 0.38375 & 0.07842 & 4.893 & [0.23, 0.54] & \textbf{< .001} \\
LLaVA 34B:noise\_level.L & -0.06578 & 0.07842 & -0.839 & [-0.22, 0.09] & .402 \\
Qwen 72B:noise\_level.L & -0.08590 & 0.07842 & -1.095 & [-0.24, 0.07] & .274 \\
Qwen 235B:noise\_level.L & -0.08731 & 0.07842 & -1.113 & [-0.24, 0.07] & .266 \\
Human:noise\_level.Q & -0.23056 & 0.07842 & -2.940 & [-0.38, -0.08] & \textbf{.003} \\
LLaVA 34B:noise\_level.Q & -0.08053 & 0.07842 & -1.027 & [-0.23, 0.07] & .305 \\
Qwen 72B:noise\_level.Q & 0.06945 & 0.07842 & 0.886 & [-0.08, 0.22] & .376 \\
Qwen 235B:noise\_level.Q & 0.08948 & 0.07842 & 1.141 & [-0.06, 0.24] & .254 \\
Human:noise\_level.C & -0.02774 & 0.07842 & -0.354 & [-0.18, 0.13] & .724 \\
LLaVA 34B:noise\_level.C & 0.12395 & 0.07842 & 1.581 & [-0.03, 0.28] & .115 \\
Qwen 72B:noise\_level.C & -0.00431 & 0.07842 & -0.055 & [-0.16, 0.15] & .956 \\
Qwen 235B:noise\_level.C & -0.02083 & 0.07842 & -0.266 & [-0.17, 0.13] & .791 \\
Human:noise\_level\textasciicircum 4 & -0.03060 & 0.07842 & -0.390 & [-0.18, 0.12] & .697 \\
LLaVA 34B:noise\_level\textasciicircum 4 & -0.01993 & 0.07842 & -0.254 & [-0.17, 0.13] & .799 \\
Qwen 72B:noise\_level\textasciicircum 4 & -0.02210 & 0.07842 & -0.282 & [-0.18, 0.13] & .778 \\
Qwen 235B:noise\_level\textasciicircum 4 & 0.02283 & 0.07842 & 0.291 & [-0.13, 0.18] & .771 \\
Human:noise\_level\textasciicircum 5 & 0.10283 & 0.07842 & 1.311 & [-0.05, 0.26] & .190 \\
LLaVA 34B:noise\_level\textasciicircum 5 & -0.00981 & 0.07842 & -0.125 & [-0.16, 0.14] & .901 \\
Qwen 72B:noise\_level\textasciicircum 5 & -0.01763 & 0.07842 & -0.225 & [-0.17, 0.14] & .822 \\
Qwen 235B:noise\_level\textasciicircum 5 & -0.03306 & 0.07842 & -0.422 & [-0.19, 0.12] & .673 \\
\bottomrule
\end{tabular}
}
\caption{Linear mixed model for \textit{entropy} in PixCC, fitted on one value per image $\times$ source cell: $\variable{entropy} \sim \variable{source} * \variable{noise\_level} + (1 \mid \variable{image})$.}
\label{tab:coef_entropy_pixcc}
\end{table}

\begin{table}[H]
\scriptsize
\setlength{\tabcolsep}{6pt}
\centering
\begin{tabular}{lrrl}
\toprule
\textbf{Effect} & $\boldsymbol{\chi^2}$ & \textbf{df} & \textbf{p} \\
\midrule
source & 4968.61 & 4 & \textbf{< .001} \\
noise\_level & 2.66 & 5 & .753 \\
source:noise\_level & 42.08 & 20 & \textbf{.003} \\
\bottomrule
\end{tabular}
\caption{Type III Wald $\chi^2$ tests for \textit{semantic head entropy} in PixCC. Linear mixed model on one value per image and source: $\variable{entropy} \sim \variable{source} * \variable{noise\_level} + (1 \mid \variable{image})$. Robustness check with humans and LLaVA 34B only: source $\chi^2(1) = 31.21$, $p < .001$; noise level $\chi^2(5) = 9.53$, $p = .090$; interaction $\chi^2(5) = 7.43$, $p = .191$.}
\label{tab:anova_entropy_pixcc}
\end{table}
 
\begin{table}[H]
\scriptsize
\setlength{\tabcolsep}{1.5pt}
\centering
\resizebox{0.5\textwidth}{!}{%
\begin{tabular}{llrrl}
\toprule
\textbf{Noise} & \textbf{Contrast} & \textbf{Estimate} & \textbf{95\% CI} & \textbf{p} \\
\midrule
0\% & Human -- LLaVA 34B & 0.170 & [-0.169, 0.509] & .647 \\
 & Human -- Qwen 72B & 2.093 & [1.754, 2.433] & \textbf{< .001} \\
 & Human -- Qwen 235B & 2.070 & [1.730, 2.409] & \textbf{< .001} \\
 & Human -- InternVL 38B & 1.957 & [1.618, 2.296] & \textbf{< .001} \\
 & LLaVA 34B -- Qwen 72B & 1.923 & [1.584, 2.263] & \textbf{< .001} \\
 & LLaVA 34B -- Qwen 235B & 1.900 & [1.560, 2.239] & \textbf{< .001} \\
 & LLaVA 34B -- InternVL 38B & 1.787 & [1.448, 2.126] & \textbf{< .001} \\
 & Qwen 72B -- Qwen 235B & -0.023 & [-0.363, 0.316] & 1.000 \\
 & Qwen 72B -- InternVL 38B & -0.136 & [-0.475, 0.203] & .808 \\
 & Qwen 235B -- InternVL 38B & -0.113 & [-0.452, 0.227] & .894 \\
\midrule
10\% & Human -- LLaVA 34B & 0.252 & [-0.087, 0.592] & .251 \\
 & Human -- Qwen 72B & 2.367 & [2.027, 2.706] & \textbf{< .001} \\
 & Human -- Qwen 235B & 2.414 & [2.074, 2.753] & \textbf{< .001} \\
 & Human -- InternVL 38B & 2.414 & [2.074, 2.753] & \textbf{< .001} \\
 & LLaVA 34B -- Qwen 72B & 2.115 & [1.775, 2.454] & \textbf{< .001} \\
 & LLaVA 34B -- Qwen 235B & 2.162 & [1.822, 2.501] & \textbf{< .001} \\
 & LLaVA 34B -- InternVL 38B & 2.162 & [1.822, 2.501] & \textbf{< .001} \\
 & Qwen 72B -- Qwen 235B & 0.047 & [-0.292, 0.386] & .996 \\
 & Qwen 72B -- InternVL 38B & 0.047 & [-0.292, 0.386] & .996 \\
 & Qwen 235B -- InternVL 38B & 0.000 & [-0.339, 0.339] & 1.000 \\
\midrule
20\% & Human -- LLaVA 34B & 0.247 & [-0.092, 0.587] & .270 \\
 & Human -- Qwen 72B & 2.412 & [2.073, 2.752] & \textbf{< .001} \\
 & Human -- Qwen 235B & 2.412 & [2.073, 2.752] & \textbf{< .001} \\
 & Human -- InternVL 38B & 2.412 & [2.073, 2.752] & \textbf{< .001} \\
 & LLaVA 34B -- Qwen 72B & 2.165 & [1.826, 2.504] & \textbf{< .001} \\
 & LLaVA 34B -- Qwen 235B & 2.165 & [1.826, 2.504] & \textbf{< .001} \\
 & LLaVA 34B -- InternVL 38B & 2.165 & [1.826, 2.504] & \textbf{< .001} \\
 & Qwen 72B -- Qwen 235B & 0.000 & [-0.339, 0.339] & 1.000 \\
 & Qwen 72B -- InternVL 38B & 0.000 & [-0.339, 0.339] & 1.000 \\
 & Qwen 235B -- InternVL 38B & 0.000 & [-0.339, 0.339] & 1.000 \\
\midrule
30\% & Human -- LLaVA 34B & 0.563 & [0.224, 0.902] & \textbf{< .001} \\
 & Human -- Qwen 72B & 2.708 & [2.369, 3.047] & \textbf{< .001} \\
 & Human -- Qwen 235B & 2.708 & [2.369, 3.047] & \textbf{< .001} \\
 & Human -- InternVL 38B & 2.708 & [2.369, 3.047] & \textbf{< .001} \\
 & LLaVA 34B -- Qwen 72B & 2.145 & [1.806, 2.484] & \textbf{< .001} \\
 & LLaVA 34B -- Qwen 235B & 2.145 & [1.806, 2.484] & \textbf{< .001} \\
 & LLaVA 34B -- InternVL 38B & 2.145 & [1.806, 2.484] & \textbf{< .001} \\
 & Qwen 72B -- Qwen 235B & 0.000 & [-0.339, 0.339] & 1.000 \\
 & Qwen 72B -- InternVL 38B & 0.000 & [-0.339, 0.339] & 1.000 \\
 & Qwen 235B -- InternVL 38B & 0.000 & [-0.339, 0.339] & 1.000 \\
\midrule
40\% & Human -- LLaVA 34B & 0.662 & [0.323, 1.002] & \textbf{< .001} \\
 & Human -- Qwen 72B & 2.725 & [2.386, 3.065] & \textbf{< .001} \\
 & Human -- Qwen 235B & 2.749 & [2.410, 3.088] & \textbf{< .001} \\
 & Human -- InternVL 38B & 2.749 & [2.410, 3.088] & \textbf{< .001} \\
 & LLaVA 34B -- Qwen 72B & 2.063 & [1.724, 2.403] & \textbf{< .001} \\
 & LLaVA 34B -- Qwen 235B & 2.087 & [1.747, 2.426] & \textbf{< .001} \\
 & LLaVA 34B -- InternVL 38B & 2.087 & [1.747, 2.426] & \textbf{< .001} \\
 & Qwen 72B -- Qwen 235B & 0.023 & [-0.316, 0.363] & 1.000 \\
 & Qwen 72B -- InternVL 38B & 0.023 & [-0.316, 0.363] & 1.000 \\
 & Qwen 235B -- InternVL 38B & 0.000 & [-0.339, 0.339] & 1.000 \\
\midrule
70\% & Human -- LLaVA 34B & 0.667 & [0.328, 1.007] & \textbf{< .001} \\
 & Human -- Qwen 72B & 2.702 & [2.362, 3.041] & \textbf{< .001} \\
 & Human -- Qwen 235B & 2.702 & [2.362, 3.041] & \textbf{< .001} \\
 & Human -- InternVL 38B & 2.702 & [2.362, 3.041] & \textbf{< .001} \\
 & LLaVA 34B -- Qwen 72B & 2.034 & [1.695, 2.374] & \textbf{< .001} \\
 & LLaVA 34B -- Qwen 235B & 2.034 & [1.695, 2.374] & \textbf{< .001} \\
 & LLaVA 34B -- InternVL 38B & 2.034 & [1.695, 2.374] & \textbf{< .001} \\
 & Qwen 72B -- Qwen 235B & 0.000 & [-0.339, 0.339] & 1.000 \\
 & Qwen 72B -- InternVL 38B & 0.000 & [-0.339, 0.339] & 1.000 \\
 & Qwen 235B -- InternVL 38B & 0.000 & [-0.339, 0.339] & 1.000 \\
\bottomrule
\end{tabular}
}
\caption{Tukey-adjusted pairwise comparisons for \textit{semantic head entropy} by noise level in PixCC. Estimates are differences in entropy (bits; first minus second source; $SE = 0.124$, $df = 551$); CIs are Tukey-adjusted.}
\label{tab:pairs_entropy_pixcc}
\end{table}


\subsection{Originality and Effectiveness Combined}\label{app:creativity}

\begin{figure}[]
    \centering
    \includegraphics[width=1\linewidth]{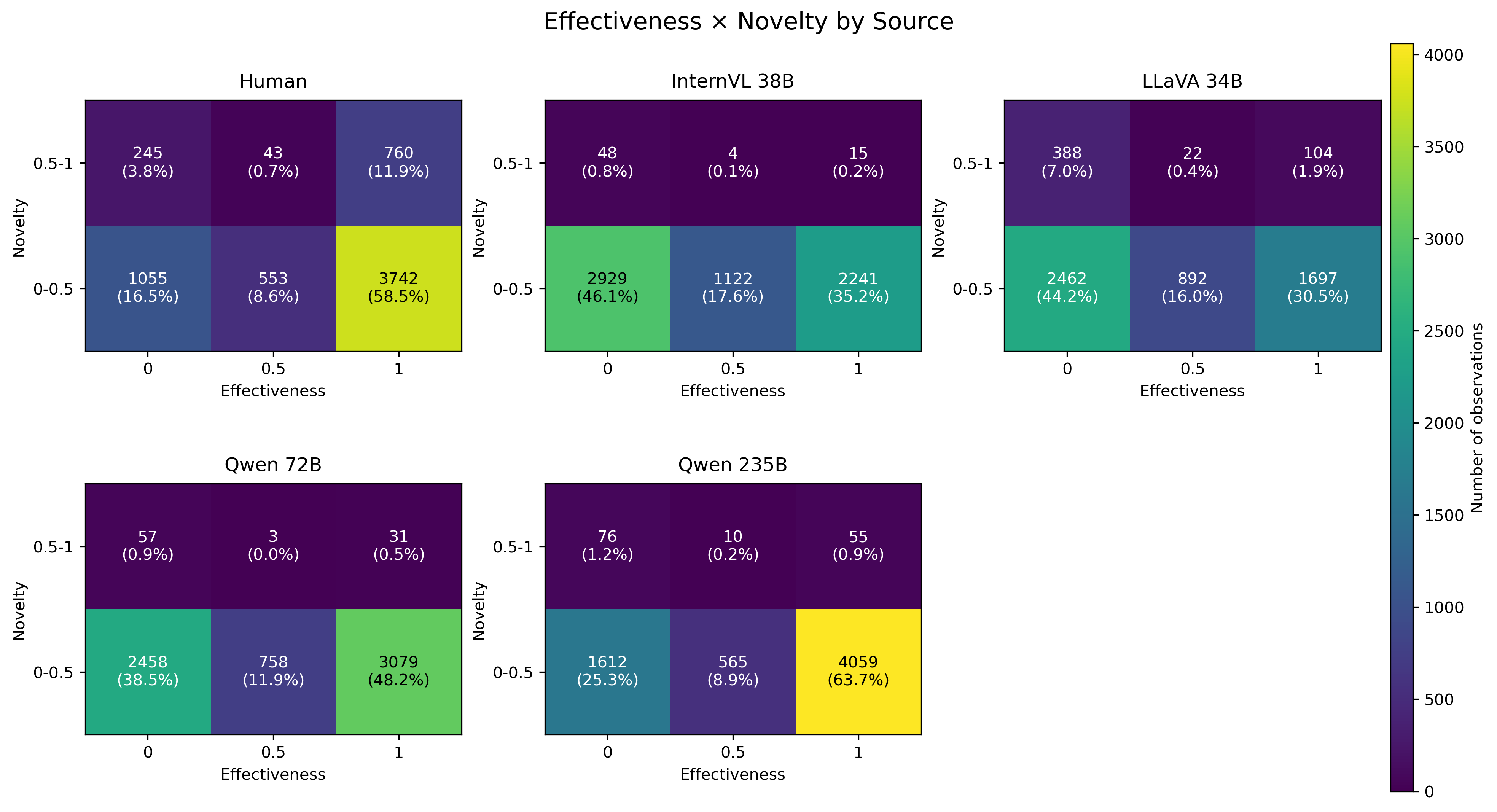}
    \caption{Effectiveness and novelty distributions across sources in SilCC}
    \label{fig:C02_effectiveness_novelty}
\end{figure}

\begin{figure}[]
    \centering
    \includegraphics[width=1\linewidth]{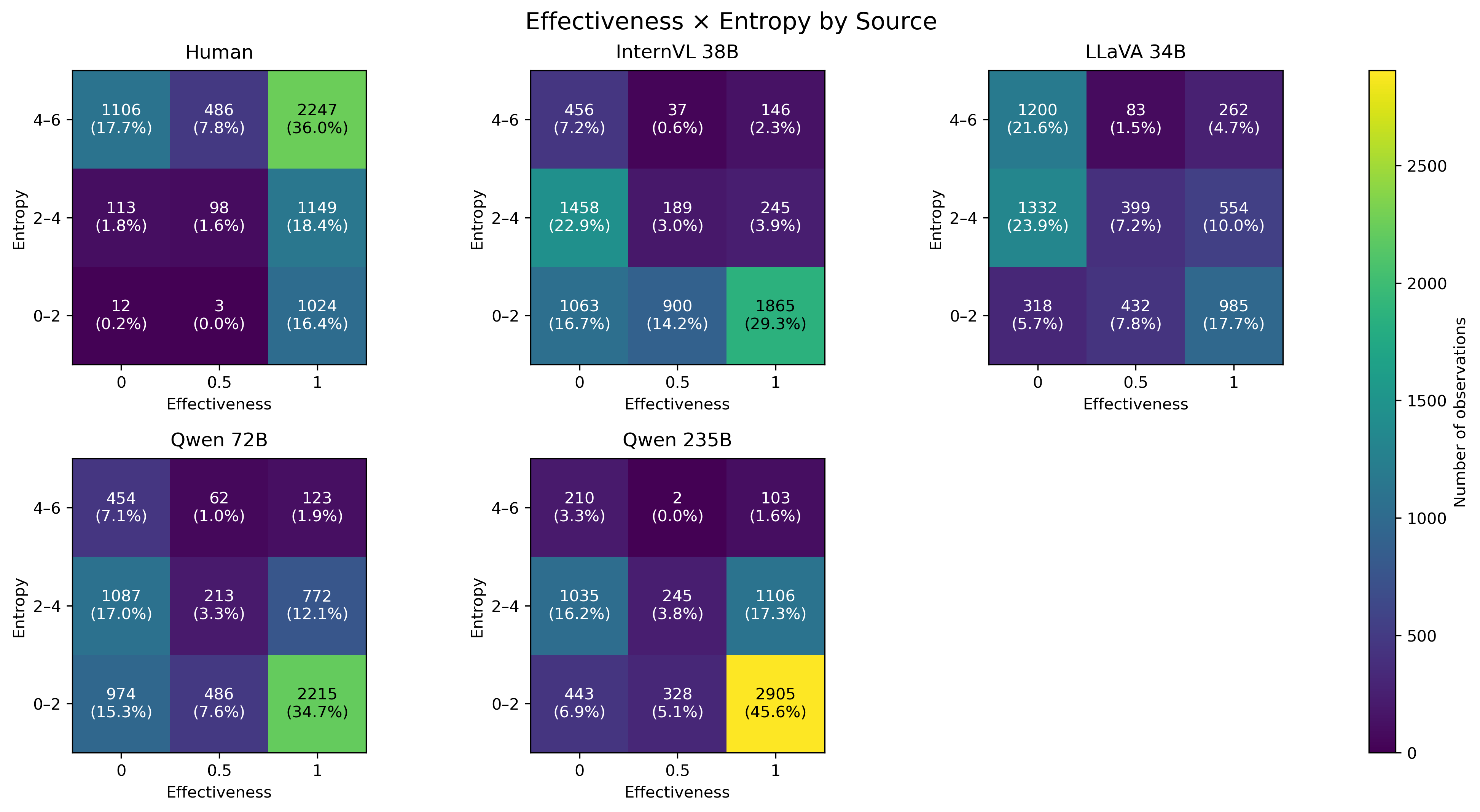}
    \caption{Effectiveness and entropy distributions across sources in SilCC}
    \label{fig:C02_effectiveness_entropy}
\end{figure}

\begin{figure}[]
    \centering
    \includegraphics[width=1\linewidth]{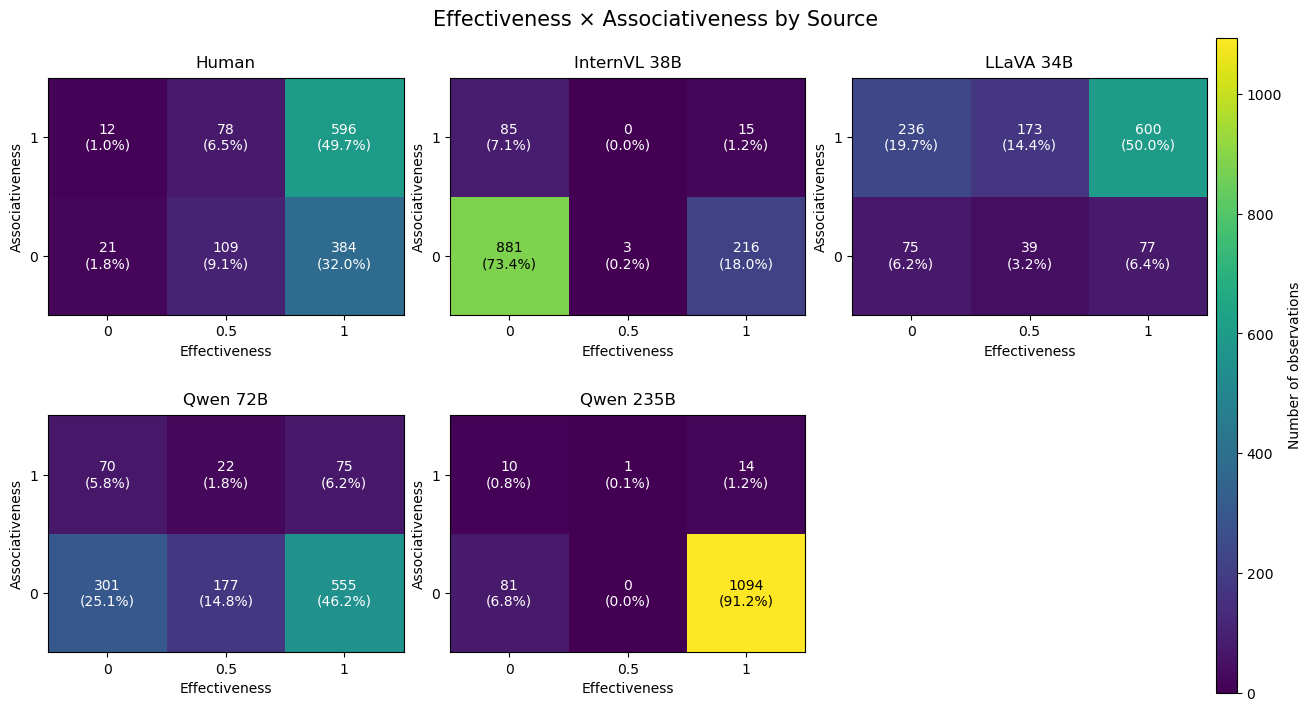}
    \caption{Effectiveness and associativness distributions across sources in PixCC}
    \label{fig:b02_effectiveness_assco}
\end{figure}

\begin{figure}[]
    \centering
    \includegraphics[width=1\linewidth]{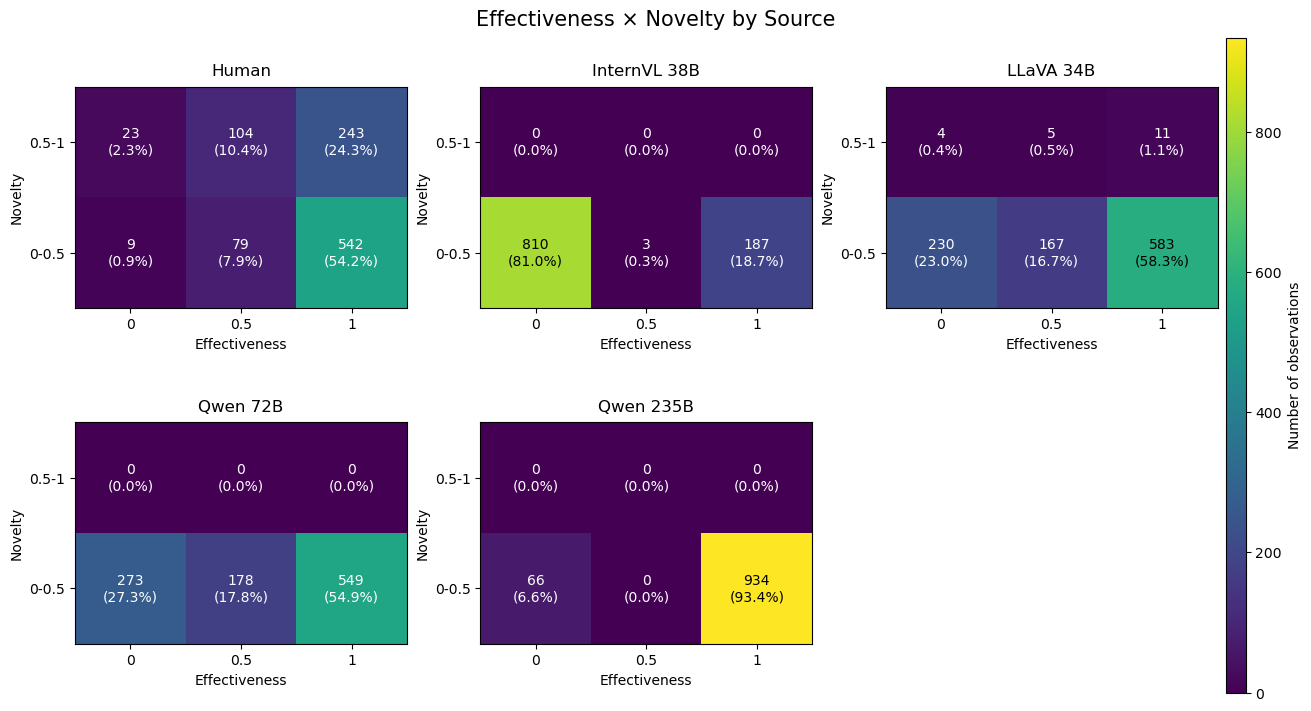}
    \caption{Effectiveness and novelty distributions across sources in PixCC}
    \label{fig:b02_effectiveness_novelty}
\end{figure}

\begin{figure}[]
    \centering
    \includegraphics[width=1\linewidth]{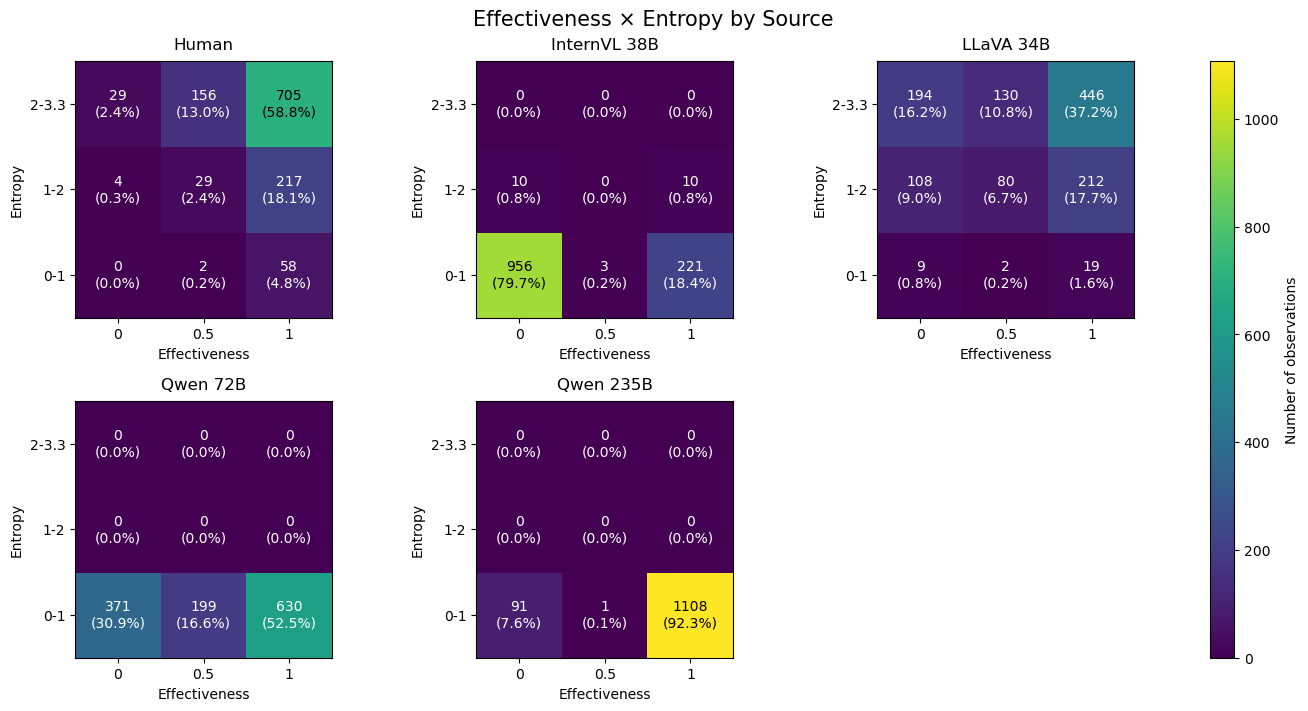}
    \caption{Effectiveness and entropy distributions across sources in PixCC}
    \label{fig:b02_effectiveness_entropy}
\end{figure}

Heatmaps (Figures~\ref{fig:C02_effectiveness_novelty}, ~\ref{fig:C02_effectiveness_entropy}, \ref{fig:b02_effectiveness_assco}, \ref{fig:b02_effectiveness_novelty} and \ref{fig:b02_effectiveness_entropy}) show the joint distribution of effectiveness (reflects) scores (0, 0.5, and 1 on x-axis) and novelty scores (0–0.5 and 0.5–1 on y-axis), the joint distribution of effectiveness scores (0, 0.5, and 1 on x-axis) and entropy bins, as well as the effectiveness scores and associativness scores (0,1) for human responses and four MLLMs. Cell annotations report the number of observations and the percentage of observations within each source. Color intensity indicates the number of observations.

\paragraph{Effectiveness $\times$ Novelty in SilCC}
Across all sources, most responses fall within the lower novelty range (0–0.5). Humans show the largest proportion of responses combining higher originality with full effectiveness, whereas higher-originality responses are comparatively rare among all four models. Among model responses in the lower originality range, Qwen 72B and especially Qwen 235B show a strong concentration of fully effective responses, while InternVL 38B and LLaVA 34B more frequently produce ineffective responses.

\paragraph{Effectiveness $\times$ Entropy in SilCC}
 Human responses show a distinct concentration of fully effective responses at both low and high entropy, whereas model distributions vary substantially by model. Larger Qwen models show a concentration of fully effective, low-entropy responses, while InternVL 38B and LLaVA 34B show more splitted distributions across effectiveness and entropy levels.

\paragraph{Effectiveness $\times$ Associatvness in PixCC}
Humans and LLaVA 34B show the largest proportion of responses combining associativity with full effectiveness. Whereas this combination is rare among the other models. Among associative responses, humans more consistently achieve full effectiveness, while LLaVA 34B also produces many partially effective or ineffective responses. Among non-associative responses, Qwen 235B shows a strong concentration of fully effective responses, whereas InternVL 38B mostly produces ineffective responses.

\paragraph{Effectiveness $\times$ Novelty in PixCC}
Humans show the highest novel and effective responses, followed by LLaVA 34B, although at a much lower rate. Among the lower-novelty responses, Qwen 235B shows a strong concentration of fully effective responses, whereas InternVL 38B mostly produces ineffective responses (The clean condition, 0\% noise, is excluded).

\paragraph{Effectiveness $\times$ Entropy in PixCC}
Humans show the highest proportion of high-entropy responses in the fully effective responses, followed by LLaVA 34B, although at a lower rate. Among the lower-entropy responses, Qwen 235B shows a strong concentration of fully effective responses, whereas InternVL 38B mostly produces ineffective responses.

\section{Implementation Details}
\label{app:implementation-details}

We evaluate several multimodal large language models, either from Hugging Face or via the Ollama runtime:
\begin{itemize}
    \itemsep0em
    \item \href{https://huggingface.co/Qwen/Qwen2.5-VL-7B-Instruct}{Qwen2.5-VL-72B-Instruct (quantized)}: temperature 0.7
    \item \href{https://ollama.com/library/qwen3-vl:235b}{Qwen3-VL-235B}: temperature 0.7.
    \item \href{https://huggingface.co/llava-hf/llava-v1.6-34b-hf}{LLaVA-v1.6-34B-HF}: temperature 0.7 
    \item \href{https://huggingface.co/OpenGVLab/InternVL3_5-38B-HF}{OpenGVLab-InternVL3\_5-38B-HF}: temperature 0.7
   
\end{itemize}

We used three NVIDIA RTX A6000 GPUs for inference with the Qwen2.5-72B model, and two GPUs of the same type for InternVL3.5-38B. Qwen3-235B-VL was run on two NVIDIA RTX PRO 6000 Blackwell Workstation Edition GPUs. LLaVA-v1.6-34B was evaluated on a single NVIDIA RTX 3090. Depending on model size, the generation of the responses took between 6 and 24 hours.

In our work, we mainly use scientific artifacts in
the form of publicly model implementations, R packages, Python frameworks and modules.

\section{Use of AI Assistants}

AI assistants were used during manuscript preparation for specific linguistic reformulation
to refine clarity and style, table formatting, to assist with code writing, and debugging the R code for the mixed-effects models. 

\end{document}